\documentclass{article}
\usepackage{arxiv}
\renewcommand{\headeright}{}
\renewcommand{\undertitle}{}
\renewcommand{\shorttitle}{MIRRA: Fast, High-Fidelity Molecular Generation}
\usepackage{etoolbox}
\makeatletter
\patchcmd{\@maketitle}{\textsc{\undertitle}\\}{}{}{\PackageWarning{manuscript}{Could not remove the preprint subtitle line}}
\makeatother
\usepackage[T1]{fontenc}
\usepackage[utf8]{inputenc}
\usepackage[authoryear,round]{natbib}
\setcitestyle{citesep={;},aysep={,},yysep={;}}
\usepackage{amsmath,amssymb,booktabs,graphicx,xcolor}
\usepackage{microtype}
\usepackage{xurl}
\usepackage[font=small]{caption}
\usepackage{hyperref}
\usepackage{placeins}
\usepackage{flafter}
\renewcommand{\floatpagefraction}{0.85}
\hypersetup{hidelinks}
\newcommand{\model}{MIRRA}
\newcommand{\TimingMemoryReduction}{46.0}
\newcommand{\TimingCurrentAllocatedGiB}{1.47}
\newcommand{\TimingFragFMAllocatedGiB}{2.73}
\newcommand{\ReleasedMirraFastSeconds}{17.09}
\newcommand{\ReleasedMirraMainSeconds}{27.83}
\newcommand{\ReleasedFragFastSeconds}{44.74}
\newcommand{\ReleasedFragBestSeconds}{78.66}
\newcommand{\ReleasedFragLongSeconds}{136.54}

\title{MIRRA: Fast, High-Fidelity Molecular Generation with
Chemistry-Informed Discrete Flow Matching}
\author{%
  \normalfont
  Ge Sun\textsuperscript{1,2}\qquad Juan J. de Pablo\textsuperscript{1,2,3}\\[5pt]
  \begin{minipage}{\dimexpr\textwidth-2\tabcolsep\relax}
  \centering\normalfont\fontsize{8}{10}\selectfont\itshape
  \textsuperscript{1}Courant Institute of Mathematical Sciences,
  New York University, New York, NY 10012\\
  \textsuperscript{2}Tandon School of Engineering,
  New York University, Brooklyn, NY 11201\\
  \textsuperscript{3}Department of Physics, New York University, New York, NY 10003
  \end{minipage}
}
\date{}
\begin{document}
\maketitle

\begin{abstract}
Fast, high-fidelity molecular generation requires accurate chemical choices and
efficient sampling. Chemical building blocks reduce graph size, but make
each node a choice among tens of thousands of related identities. We introduce
\model{} (Molecular Identity Representation, Refresh and Assembly), a discrete
flow-matching generator that learns from chemical knowledge and evaluates
identity probabilities only when needed, while keeping every identity in its vocabulary
available. Chemical features share information across related blocks, and
learned identity corrections distinguish them in molecular context. By deciding
which positions will refresh before computing their probabilities, the sampler
avoids unused predictions while preserving each sampling transition. Against
the strongest baselines in each task, \model{} reduces chemical and graph
distribution errors on ZINC250k by 29--31\% and 53--60\%, respectively, with
generation up to 8.6 times faster. Property targeting yields an estimated
7.8--20.8 times more distinct novel hits per second. On a held-out polymer test set,
\model{} achieves 100\% periodic validity with known chain topology, versus 57.0\% for
a polymer variational autoencoder. These results demonstrate fast,
high-fidelity generation with the full chemical vocabulary available throughout
sampling.
\end{abstract}

\section{Introduction}

Molecules reuse chemistry. Rings and functional groups recur across compounds,
so selecting one building block can specify several atoms and bonds at once.
This makes substructure-based generation attractive: it reduces the number of
graph variables while preserving chemically meaningful units
\citep{brics,jtnn,han2025functional,fragfm}. Compression of the graph, however,
relocates the difficulty to the individual nodes. Instead of choosing among a
few atom types, a model must predict an identity from tens of thousands of
building blocks, each in the context of a different partially formed molecule.

Two requirements follow. First, the predicted probabilities must reflect both
chemical similarity and individual difference. Two aromatic blocks may share
useful structural information while differing in ring size, heteroatoms or
attachment pattern, and these differences can determine which surrounding
structures they fit. Second, evaluating those probabilities must remain
affordable, since the vocabulary is scored at every position of every sampling
step; a large vocabulary can otherwise erase the computational benefit of the
smaller graph.

\begin{figure}[t]
\centering
\includegraphics[width=\linewidth,height=0.52\textheight,keepaspectratio]{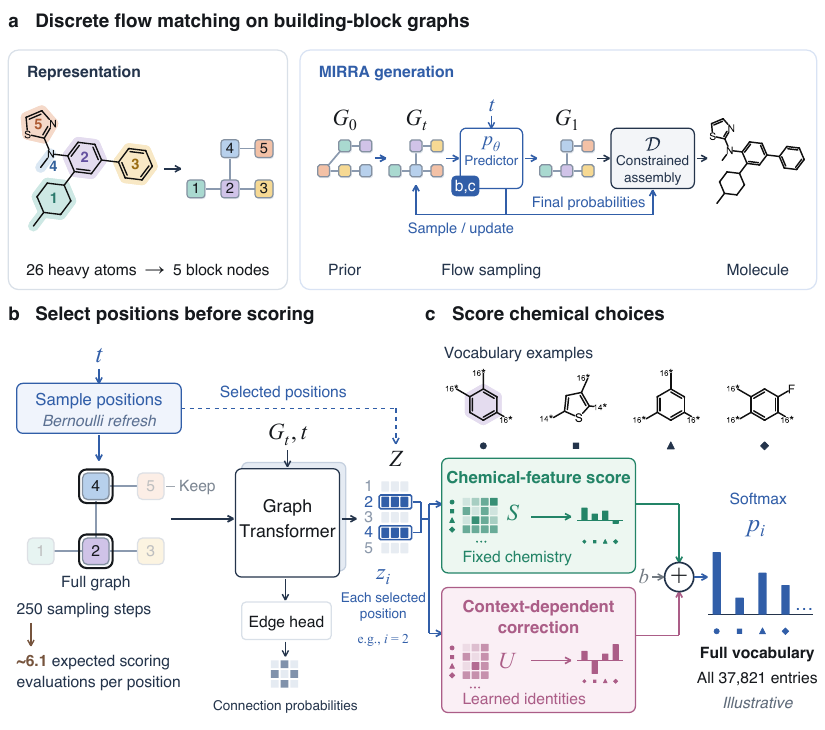}
\caption{\textbf{MIRRA combines chemistry-informed prediction, refresh scheduling
and constrained assembly.}
(a) A 26-heavy-atom molecule becomes five building-block nodes. The predictor
$p_\theta$ supplies node and connection probabilities for discrete flow matching,
using refresh scheduling (b) and chemistry-informed identity scores (c).
Constrained assembly $\mathcal{D}$ uses final identities and these probabilities
to construct a tree over the blocks and pair compatible attachment sites, with
limited identity changes when needed.
(b) Independent Bernoulli refresh decisions select positions for vocabulary
scoring using the full graph as context. A selected identity can remain
unchanged. At 250 steps, expected scoring evaluations per position fall from 250 to 6.1. (c) Chemical-feature and
identity-correction scores, plus a frequency offset, define probabilities over
all 37,821 entries. The four chemical examples are vocabulary members, not a
shortlist; intermediate states, matrices and probability bars are schematic.}
\label{fig:short-model}
\end{figure}

We ask: \emph{can a generator retain detailed chemical choices and still be
fast?} \model{} (Molecular Identity Representation, Refresh and Assembly)
answers by combining chemistry-informed prediction with refresh scheduling
(Figure~\ref{fig:short-model}). Every identity in its vocabulary remains
available throughout generation. The \emph{representation} assigns each block a
probability from two terms: fixed chemical features shared across related
entries, and learned per-entry corrections that depend on molecular context.
The \emph{refresh} schedule fixes which positions will be resampled before any
scoring occurs, so only those positions evaluate the full vocabulary while the
complete graph still supplies context. \emph{Assembly} converts the sampled
building blocks into a molecule by selecting a tree of connections and
resolving compatible attachment sites from the terminal probabilities.

\newpage
Our contributions are:
\begin{itemize}
\setlength{\itemsep}{3pt}
\setlength{\parsep}{0pt}
\interlinepenalty=10000
\item \textbf{Higher fidelity at lower cost.} \model{} achieves the best chemical
and graph fidelity in our ZINC250k evaluation. Compared with the strongest
baseline, FragFM at 500 steps, \model{} at 25 steps lowers Fr\'echet ChemNet
Distance (FCD) by 28.7\% and generates 8.6 times faster. Both models use the
same training data, and matched timings include assembly.
\item \textbf{Chemical sharing and identity corrections improve fidelity.}
Chemical features share learning across related identities, while contextual
corrections preserve their differences. Matched training controls and
generation/assembly interventions establish how both components improve fidelity.
\item \textbf{Faster sampling through prediction-independent scheduling.}
Prediction-independent refresh scheduling computes identity probabilities only
for positions selected to refresh, scoring every vocabulary entry at each
refresh. It preserves the finite-step kernel and gives a 3.0-fold speedup at
fixed weights.
\item \textbf{Beyond the benchmark.} \textbf{Conditional design} increases
estimated distinct novel target hits per second. On a reserved \textbf{polymer}
test, \model{} achieves
\textbf{100\% periodic validity} under known chain topology. The fixed design
also transfers to MOSES.
\end{itemize}

\newpage
\section{Related work}
\label{sec:related}

\subsection{Molecular generation}
GDSS, DiGress, GruM and SID generate graphs through diffusion and iterative
denoising \citep{gdss,digress,grum,sid}. SyCo maps molecular graphs to synthetic
Euclidean coordinates and generates in that latent space \citep{syco}.
GraphBFN evolves continuous beliefs over graph categories \citep{graphbfn},
while HDFM and HOG-Diff organize generation coarse-to-fine \citep{hdfm,hogdiff}.
For property control, STGG+ combines conditional spanning-tree
generation with self-criticism \citep{stggplus}, while TreeDiff guides diffusion
through inference-time tree search \citep{treediff}.
\model{} builds on discrete flow matching \citep{dfm}, with
chemically structured output sharing and refresh scheduling that preserves
the specified finite-step transition. For conditional generation, it measures
useful molecular yield as distinct novel target hits per unit time.

Fragment-based generators pair compact molecular graphs with assembly.
The junction-tree VAE generates substructure scaffolds and assembles molecules,
while HierVAE predicts larger motifs and their attachments \citep{jtnn,hiervae}.
Functional-group coarse-graining provides chemically meaningful graph nodes
\citep{han2025functional}, and NSGGM combines neural scaffold proposals with
symbolic constraint solving for final assembly \citep{nsggm}.
FragFM learns fragment embeddings, predicts
identities through contrastive fragment bags, and resolves attachments by
constrained matching \citep{fragfm}. SynFormer generates synthetic pathways
and selects building blocks by generating molecular fingerprints and retrieving
nearby entries from a database \citep{synformer}. \model{} assigns an explicit
probability to every fragment identity at each refresh and uses terminal
probabilities to select a connection tree and compatible attachment pairs.
Section~\ref{sec:vocab-cost} compares full-vocabulary execution with a
prior-sampled candidate subset, isolating how restricting the available
identities affects fidelity and cost.

Sharing information across output identities provides a complementary approach.
Label Embedding Network learns relationships between labels
\citep{label_embedding}; deep residual output layers share learned label mappings
in language generation \citep{drill}; and LGSA aligns hierarchical label geometry
for text classification \citep{lgsa}. \model{} combines fixed chemical output
features with context-dependent identity corrections and trains probabilities
over the full vocabulary. Chemical structure shares learning across molecular
identities, while learned corrections express their differences in context.
Matched training controls test chemical alignment, and generation/assembly
interventions identify how the corrections improve fidelity.

\subsection{Efficient discrete sampling}
Structured discrete diffusion supports categorical and absorbing-state
corruption \citep{d3pm}. MELD learns element-specific corruption schedules to
reduce collisions between molecular diffusion trajectories \citep{meld},
while Categorical Flow Maps distill flows for generation in fewer steps
\citep{categorical_flow_maps}. MDLM and RADD accelerate absorbing
diffusion by caching time-independent network outputs while the partially
revealed sequence remains unchanged \citep{mdlm,radd}. First-hitting samplers
directly sample unmasking times \citep{first_hitting}; EDGE and DeFoG study
sparse active-node and adaptive graph sampling \citep{edge,defog}.
DDPD separates a learned planner that selects positions from a denoiser that
predicts their values \citep{ddpd}. \model{} selects refresh positions from
prediction-independent Bernoulli gates, allowing this decision to precede
identity scoring while preserving the finite-step kernel. Identities can refresh
repeatedly, and full-vocabulary probabilities are computed only at selected
positions. The time-conditioned backbone still processes each active graph,
so the main saving comes from avoiding unused vocabulary projections.
The harmonic count $H_T$ measures per-node refresh evaluations;
absorbing-diffusion caching counts whole-network evaluations between changes
to the sequence.

\subsection{Polymer generation}
BigSMILES describes stochastic macromolecular structures \citep{bigsmiles}, and
G$^2$RINS provides a generative string-and-graph representation of hierarchical
polymers \citep{g2rins}. HiPoly builds on G$^2$RINS for polymer property
prediction and generative design \citep{hipoly}. The Open Macromolecular Genome links repeat-unit
generation to polymerization chemistry \citep{omg}, while Graph DiT studies
multi-property-conditioned polymer generation \citep{graph_dit}.
\model{} extends chemically structured prediction to periodic repeat units and
evaluates structural fidelity and periodic validity on a reserved polymer split
with known chain topology.

\clearpage

\section{\model{}: representation, refresh and assembly}
\label{sec:model}

\subsection{Representation: chemically structured identities}
A molecule is represented by building-block (molecular fragment) identities
$X=(x_1,\ldots,x_n)$ and a symmetric binary adjacency matrix $E$. Blocks expose typed attachment sites
obtained from BRICS-derived decompositions \citep{brics}, while their identities
specify internal atoms and bonds. The generator therefore predicts only the
block graph. Distinct substitution and attachment patterns remain distinct
choices. The benchmark vocabulary contains 37,821 entries; its finite size also
bounds the chemical space this representation expresses.

Each entry $v$ has a fixed vector $s_v$ describing chemical structure and
attachments. A graph transformer reads these features, the noisy edges and the
time to produce a contextual state $z_i$ for each node, and the model scores
every entry as
\begin{equation}
\ell_i(v)=a\,s_v^\top g(z_i)+u_v^\top Dz_i+b_v,
\qquad p_i(v)=\operatorname{softmax}_v\ell_i(v).
\label{eq:head}
\end{equation}
The first term projects the contextual state into feature space through $g$
with learned scale $a$, allowing related blocks to inform one another's
learning. The second is the identity correction, a learned vector $u_v$ for
each entry read through a shared low-rank projection $D$. The fixed offset
$b_v$ encodes training frequency. The model is trained with cross-entropy over
probabilities normalized across the full vocabulary.

The two terms serve complementary purposes. For two entries with identical
chemical features, the relative odds are fixed across all molecules without the
correction; with it, their log-odds differ by $(u_v-u_w)^\top Dz_i$ and depend
on context, so ring similarity does not force two attachment patterns to be
interchangeable. This is a statistical use of chemical structure rather than an
energy or reaction model: it expands output expressivity without recovering
distinctions absent from the input features
(Appendix~\ref{app:mechanism-algebra}).

Training uses the linear independent-coupling path of discrete flow matching
\citep{dfm}: given clean identity $x_1$ and source prior $p_0$,
\begin{equation}
q_t(x_t\mid x_1)=t\,\delta_{x_1}(x_t)+(1-t)p_0(x_t),
\label{eq:path}
\end{equation}
with an analogous symmetric binary corruption on edges. Individual identities
remain explicit in the state throughout; chemical sharing changes their predictor,
not the state. Appendix~\ref{app:model-details} specifies the features,
dimensions and training recipe.

\subsection{Refresh: scoring only at scheduled positions}
For $T$ sampling steps, we use the grid $t_k=k/T$, $k=0,\ldots,T-1$. The finite-step node transition is
\begin{equation}
K_{i,k}(v\mid X,E)=(1-\alpha_k)\mathbf{1}[v=x_i]
       +\alpha_k\,p_{\theta,i}(v\mid X,E,t_k),\qquad
\alpha_k=\frac{1}{T-k}.
\label{eq:refresh-kernel}
\end{equation}
The transition is a mixture of retaining the current identity and drawing a new
one from the model, and a conventional implementation evaluates $p_{\theta,i}$
before sampling from it. The mixture weight $\alpha_k$, however, is fixed by
the schedule and independent of the prediction, so the two operations may be
reordered and the Bernoulli refresh drawn first. When the refresh is zero the
identity is retained, and its vocabulary probabilities cannot affect the
transition, so they need not be computed; when it is one, all 37,821 entries
are scored and an identity is sampled, which may coincide with the current one.
Edges are governed by analogous gates, all predictions use the incoming graph,
and updates are simultaneous.

Because the gate is independent of the prediction, this ordering preserves
Equation~\ref{eq:refresh-kernel} in real arithmetic, and expected
full-vocabulary projections per node decrease from $T$ to
$H_T=\sum_{j=1}^{T}1/j$, approximately 6.1 instead of 250. Measured refresh counts agree with this prediction (Appendix~\ref{app:refresh-counts}). Early refreshes are
sparse and the final step refreshes every node, while a graph in which nothing
refreshes requires no forward pass. Within each active graph, the backbone and
edge predictor remain dense; end-to-end speed is measured separately.

\subsection{Assembly: from block graph to molecule}
\label{sec:assembly}
The assembly decoder converts the sampled building blocks into a connected
molecule. It reuses the node and edge probabilities computed at the final
sampling transition without a new prediction, and selects connections from
the terminal edge probabilities rather than the sampled adjacency. The decoder
joins the blocks into a tree and uses every attachment site exactly once;
joined sites must also have compatible types. Let $\tilde{x}_i$ denote the
identity at position $i$ after any decoder changes, $m(\tilde{x}_i)$ its
attachment count, and $d_i$ its degree in the tree. These requirements give:
\begin{equation}
d_i=m(\tilde{x}_i),
\qquad\text{and hence}\qquad
\sum_{i=1}^{n}m(\tilde{x}_i)=2(n-1).
\label{eq:assembly-constraints}
\end{equation}

Before constructing a tree, the decoder checks the total-site condition on the
right. For $n>1$, every block must also have at least one attachment site. If
these count requirements are unmet, node probabilities guide changes to the
sampled identities. Once the counts are feasible, terminal edge probabilities
guide a bounded search over degree-exact trees. Training-derived attachment
preferences rank the compatible site pairings of each candidate. If this
search finds no valid molecule, the decoder tries a bounded set of identity
replacements that preserve the attachment counts. The search and repair
procedures are specified in Appendix~\ref{app:assembly}.

Section~\ref{sec:controls} separately examines identity correction during
generation and terminal assembly. Failed assemblies count as failed generation
attempts and are not replaced.

\section{High fidelity at low sampling cost}
\label{sec:benchmark}

\paragraph{Evaluation protocol.}
We measure chemical distribution matching with FCD \citep{fcd}, graph discrepancy with NSPDK MMD$^2$ \citep{gdss}, and validity,
uniqueness and novelty separately. For results evaluated in this study in
Table~\ref{tab:short-published} and Figures~\ref{fig:short-frontier}--\ref{fig:short-stages},
we score valid molecules from 10,000 attempts against all 24,887 ZINC
references; failures are not replaced. ZINC informed architecture and correction-rank
selection, so MOSES and the reserved polymer test evaluate the fixed design
(Appendix~\ref{app:current-protocol}).

\begin{table}[!htbp]
\caption{\textbf{High fidelity at few sampling steps on ZINC250k.}
Both MIRRA settings lead on FCD and NSPDK. \textbf{Bold}/\underline{underline}
mark best/second-best among compared models, using unrounded values.}
\label{tab:short-published}
\centering
\small
\begin{tabular*}{\linewidth}{@{\extracolsep{\fill}}lrrr@{}}
\toprule
\multicolumn{4}{@{}l}{\textbf{(a) Fidelity and validity: all baselines}}\\
Model (sampling steps) & FCD $\downarrow$ & NSPDK $\times10^3$ $\downarrow$ & Valid (\%) $\uparrow$\\
\midrule
Training reference & 0.336 & 0.110 & n/a\\
\midrule
GDSS (1000) & 14.656 & 19.500 & 97.01\\
DiGress (1000) & 3.482 & 2.100 & 94.99\\
GruM (1000) & 2.257 & 1.500 & 98.65\\
GraphBFN (500) & 2.116 & 1.300 & 99.22\\
SID, marginal (500) & 2.010 & 2.060 & \underline{99.50}\\
DFM (128 NFE) & 1.480 & 0.554 & 99.41\\
DeFoG (500) & 1.425 & 0.800 & 99.22\\
HDFM (128 NFE) & 1.230 & 0.433 & 99.30\\
FragFM (500) & 0.528 & 0.382 & 99.14\\
\midrule
MIRRA (25) & \underline{0.377} & \underline{0.180} & 99.27\\
MIRRA (250) & \textbf{0.366} & \textbf{0.154} & \textbf{99.81}\\
\bottomrule
\end{tabular*}
\par\smallskip
\begin{minipage}[t]{0.48\linewidth}
\begin{tabular*}{\linewidth}[t]{@{\extracolsep{\fill}}lr@{}}
\multicolumn{2}{@{}l}{\textbf{(b) Uniqueness}}\\
Model (steps) & Unique (\%) $\uparrow$\\
\midrule
GDSS (1000) & 99.64\\
DiGress (1000) & 99.97\\
GruM (1000) & 99.97\\
SID (500) & 99.84\\
FragFM (500) & \underline{99.98}\\
\midrule
MIRRA (25) & \textbf{99.98}\\
MIRRA (250) & 99.97\\
\bottomrule
\end{tabular*}
\end{minipage}\hfill
\begin{minipage}[t]{0.48\linewidth}
\begin{tabular*}{\linewidth}[t]{@{\extracolsep{\fill}}lr@{}}
\multicolumn{2}{@{}l}{\textbf{(c) Novelty}}\\
Model (steps) & Novel (\%) $\uparrow$\\
\midrule
GDSS (1000) & \textbf{100.00}\\
DiGress (1000) & \underline{99.99}\\
GruM (1000) & 99.98\\
SID (500) & 99.97\\
FragFM (500) & 99.23\\
\midrule
MIRRA (25) & 97.69\\
MIRRA (250) & 97.61\\
\bottomrule
\end{tabular*}
\end{minipage}
\par\smallskip
\begin{minipage}{\linewidth}\footnotesize
Results evaluated in this study use 10,000 attempts and 24,887 references; MIRRA250 is a three-draw
per-metric median, other results evaluated here use one draw. Training reference is not ranked;
validity is not applicable. Diversity panels use the same models with both
metrics available. Sources and protocols: Appendix~\ref{app:current-protocol}.
\end{minipage}
\end{table}

\subsection{The fidelity advantage survives few sampling steps}
\begin{figure}[!htbp]
\centering
\includegraphics[width=0.85\linewidth]{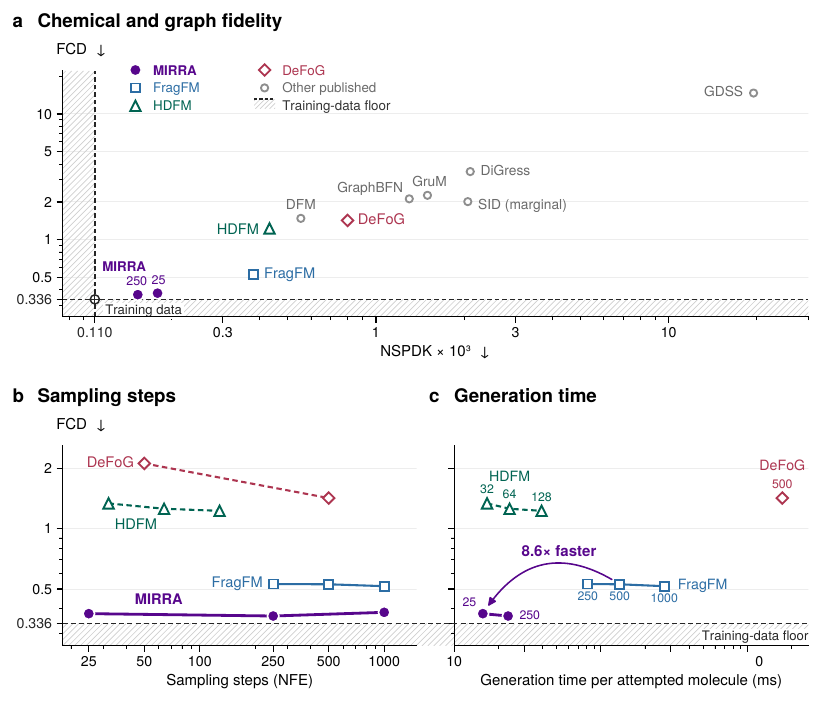}
\caption{\textbf{Strong chemical and graph fidelity at low sampling cost.}
(a) FCD and NSPDK discrepancies; (b) FCD versus steps (reported number of function evaluations, NFE, for HDFM);
(c) FCD versus time per attempted molecule. Lower is better. Quality for MIRRA
and FragFM uses 10,000 attempts; MIRRA250 is a three-draw median. MIRRA
and FragFM suffixes denote sampling steps. Hatching marks the
training-data floor, measured from a fixed training sample. The 8.6-fold arrow is the
matched MIRRA25--FragFM500 timing comparison; their times are medians of four
runs (three for FragFM250/1000). See Appendix~\ref{app:current-protocol}.}
\label{fig:short-frontier}
\end{figure}

At 250 steps, \model{} reaches FCD \textbf{0.366} versus 0.528 for FragFM at
500 steps, a \textbf{30.6\%} reduction under the same evaluation protocol, and
NSPDK$\times10^3$ improves from 0.382 to 0.154. At 25 steps, FCD remains 28.7\%
below FragFM500 while matched generation is \textbf{8.6 times faster},
including assembly (Figure~\ref{fig:short-frontier}). These are complete
generators trained on the same data, though capacities, representations and
recipes differ.

The broader comparison places both \model{} settings below every baseline in Table~\ref{tab:short-published} on FCD and NSPDK, and MIRRA250 has
the highest validity in the table at 99.81\%, with 99.97\% uniqueness and 97.61\%
novelty.

\subsection{Scheduling, not restriction, makes the vocabulary practical}
\label{sec:vocab-cost}
We run the benchmark checkpoint in three ways at 250 steps: score every entry
at every position; score every entry only at scheduled positions; or score only
a candidate subset at scheduled positions. The subset contains 1,536 distinct
prior-sampled entries plus the current identities, renormalized over that set,
and its runtime includes candidate construction. All three modes supply
assembly with the same complete terminal scores.

\begin{figure}[!htbp]
\centering
\includegraphics[width=0.88\linewidth]{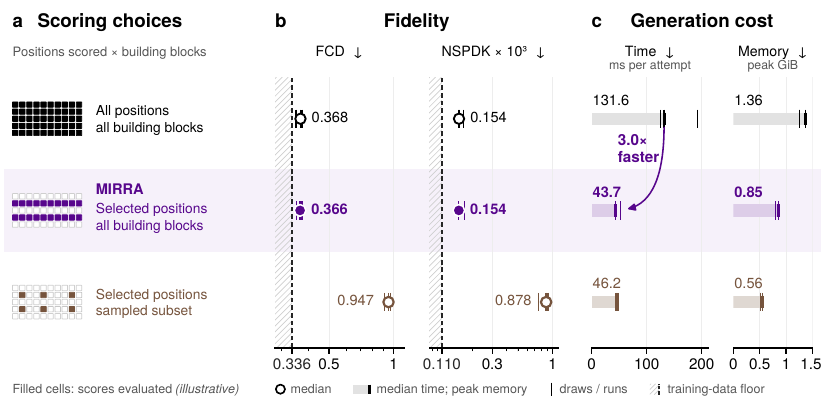}
\caption{\textbf{Schedule the computation while preserving the choices.}
(a) The positions and entries scored by three implementations of one checkpoint.
(b) FCD and NSPDK: circles are three-draw medians; ticks show every draw.
(c) Median time over three separate 1,024-attempt runs and peak allocated tensor
memory, with individual runs shown. Scheduling is 3.0 times faster than dense
scoring and slightly faster than the tested subset. Quality follows the
10,000-attempt protocol of Figure~\ref{fig:short-frontier}; timings use the same GPU and batch size 128. Candidate restriction has no separate training
or proposal correction, and variation reflects sampling and timing repetitions.}
\label{fig:short-vocabulary}
\end{figure}

Scheduling reduces median time from 131.6 to 43.7 ms per attempt, a
\textbf{3.0-fold speedup}, with nearly identical observed FCD (0.3680 versus
0.3665). The candidate subset behaves differently: its FCD rises to 0.9469 and
NSPDK$\times10^3$ to 0.8780, and at 46.2 ms it is marginally slower than
scoring all 37,821 entries under the same schedule
(Figure~\ref{fig:short-vocabulary}), while saving 34.7\% of peak allocated
tensor memory. For this implementation and checkpoint, restricting the
choices therefore yields memory savings rather than speed, at a substantial
cost in fidelity (Appendix~\ref{app:vocabulary-cost}).

\paragraph{Keeping the full vocabulary preserves chemical fidelity.}
Candidate restriction can preserve almost all valid outputs
while changing molecular composition: in the three-model controls, paired comparisons reveal a molecular-size shift before assembly. An elementary argument
explains why. Even if exact predictions preserve the target distribution,
proposing one candidate from a nonuniform prior and renormalizing over it and
the current identity can overrepresent common entries, and smaller time steps
need not remove that bias (Appendix~\ref{app:support}).

\section{Why chemical sharing and identity corrections work together}
\label{sec:controls}
\begin{table}[!htbp]
\caption{\textbf{Correct chemical features improve both chemical and graph
fidelity on ZINC250k.} Independently trained predictors share the input
features, backbone, training budget and assembly rule, with comparable capacity.
All rows use 250 sampling steps and the common final training update.
Best values are bold; second-best values are underlined.}
\label{tab:short-controls}
\centering
\begingroup
\small
\setlength{\tabcolsep}{3pt}
\begin{tabular*}{\linewidth}{@{\extracolsep{\fill}}lrrr@{}}
\toprule
Output predictor & FCD $\downarrow$ & NSPDK $\downarrow$ & Novel (\%) $\uparrow$ \\
& & $(\times10^3)$ & \\
\midrule
Conventional learned predictor & 0.1793 & 0.1223 & \textbf{98.05} \\
Shuffled chemical features & 0.1757 & 0.1547 & 97.55 \\
Chemistry shuffled within matched groups & \underline{0.1655} & \underline{0.1188} & 97.51 \\
\midrule
MIRRA: correctly assigned chemical features & \textbf{0.1561} & \textbf{0.1063} & \underline{97.81} \\
\bottomrule
\end{tabular*}
\par\smallskip
\begin{minipage}{\linewidth}\footnotesize
Each row is one independently trained model at the common final update.
Matched groups preserve attachment types and frequency bins. Correct alignment improves FCD by \textbf{11--16\%} across three independently trained pairs. FCD and NSPDK use 25,000 valid outputs and reference molecules; novelty uses unique valid outputs.
\end{minipage}
\endgroup
\end{table}

\paragraph{Chemical alignment improves what the model learns.}
Chemical features could in principle help only by adding parameters. To
separate chemical information from parameter count, we independently train
predictors with correct or shuffled output-feature assignments, holding input
features, individual labels, capacity and the training recipe fixed.
Correct alignment improves FCD by \textbf{11--16\% across three matched
training pairs}. A narrower shuffle within attachment-type and frequency-bin
groups also worsens fidelity, the more informative outcome: the benefit follows
not from a structured output space in general, but from one that is chemically
correct. At a common final update, \model{} outperforms both shuffles and a
conventional learned predictor on FCD and NSPDK
(Table~\ref{tab:short-controls}), while that predictor has slightly higher
novelty (Appendix~\ref{app:design-details}).

\begin{figure}[!htbp]
\centering
\includegraphics[width=0.90\linewidth]{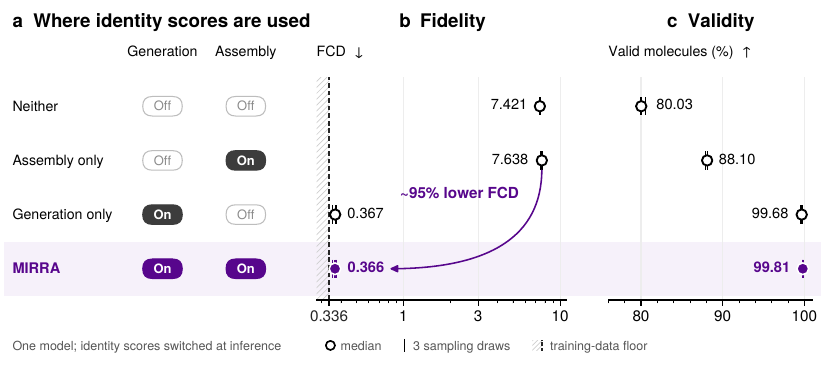}
\caption{\textbf{Assembly cannot replace accurate identity choices during generation.}
(a) A fixed-weight intervention includes or omits identity corrections during generation and assembly.
(b) FCD on a logarithmic axis; (c) validity with an axis starting at 76\%.
Circles are medians and ticks show three sampling draws of the benchmark
checkpoint. Assembly conditions within a generation arm share terminal states
and edge probabilities. The full condition is exactly the MIRRA250 population
in Figures~\ref{fig:short-frontier}--\ref{fig:short-vocabulary}; all use the same
reference and evaluation. The roughly 95\% FCD reduction compares full MIRRA
with assembly-only correction. }
\label{fig:short-stages}
\end{figure}

\paragraph{Identity corrections guide the generation trajectory.}
The corrections could in principle act only at the end, where assembly reads
the terminal probabilities. We therefore enable them independently during
generation and during terminal assembly (Figure~\ref{fig:short-stages}).
Holding full assembly scores fixed, enabling generation corrections changes
median FCD from 7.638 to 0.366, with paired reductions of 95.1--95.4\%.
Assembly-only corrections cannot recover this loss: they raise validity from
80.03\% to 88.10\% while FCD remains near 7.6. The model therefore uses
contextual identity preferences during generation. These interventions show
how the trained model uses its corrections; Table~\ref{tab:short-controls}
compares independently trained predictors.

\section{What fast, faithful generation enables}
\label{sec:applications}

\paragraph{Transfer beyond the development benchmark.}
On the official MOSES scaffold-test reference \citep{moses}, the fixed design
shows a mean FCD advantage over matched shuffled and conventional predictors,
though the effect is smaller than on ZINC, varies across training runs and step counts, and
comes with lower novelty (Appendix~\ref{app:moses-training}).

\begin{minipage}{\linewidth}
\paragraph{Valid periodic structures on a reserved polymer test.}
A split grouped by periodic identity reserves 9,949 test identities, with
checkpoints and generated samples fixed before test evaluation. Every sampled
\model{} repeat unit is periodically valid, versus 57.03\% for OMG-VAE
\citep{omg} and 35.49\% for Graph DiT (Table~\ref{tab:short-polymer}); periodic
validity is stricter than parseability, requiring a connected repeat unit, two
attachment ports and successful periodic canonicalization. At 250 steps the
structural discrepancy is 0.021 in MMD$^2\times10^3$, with 98.30\% novelty.
\model{} receives known linear chain topology while baselines generate full
encodings, and variability is measured across two sampling runs per checkpoint (Appendix~\ref{app:polymer-training}).

\end{minipage}

\begin{table}[!htbp]
\caption{\textbf{Valid polymer repeat units with closer structural distributions.}
Evaluation uses 9,949 reserved periodic identities. MIRRA has 100\% periodic
validity at both step counts. Best/second-best values are bold/underlined; ties share rank.}
\label{tab:short-polymer}
\centering
\medskip
\begingroup
\small
\setlength{\tabcolsep}{3pt}
\begin{tabular*}{\linewidth}{@{\extracolsep{\fill}}llrrrr@{}}
\toprule
Model & Steps & Structural distance $\downarrow$ & Valid (\%) & Unique (\%) & Novel (\%) \\
& & (MMD$^2\times10^3$) & $\uparrow$ & $\uparrow$ & $\uparrow$ \\
\midrule
OMG-VAE & Native & $30.498\pm0.458$ & \underline{57.03} & 97.32 & \textbf{100.00} \\
Graph DiT & 500 & $101.325\pm2.048$ & 35.49 & \textbf{100.00} & \textbf{100.00} \\
\midrule
MIRRA & 500 & $\underline{0.079\pm0.030}$ & \textbf{100.00} & \underline{99.96} & \underline{98.57} \\
MIRRA & 250 & $\mathbf{0.021\pm0.015}$ & \textbf{100.00} & 99.91 & 98.30 \\
\bottomrule
\end{tabular*}
\par\smallskip
\begin{minipage}{\linewidth}\footnotesize
Means over two 4,096-attempt draws of one checkpoint per method; distance
uncertainty is sample SD. Scorable fractions are 56.10\% for OMG-VAE,
35.49\% for Graph DiT and 100\% for MIRRA; distances condition on these
outputs. Native denotes autoregressive decoding.
\end{minipage}
\endgroup
\end{table}

\paragraph{More useful property-targeted outputs per unit time.}
We target drug-likeness (QED)~\citep{qed} and lipophilicity (logP) using
small property-dependent adapters that modify transformer states while
the benchmark backbone stays frozen. All methods use
500 steps and validation-selected guidance at the 10th, 50th and 90th training
quantiles. A useful output is valid, lies within the target window and has a
distinct identity absent from training; its estimated rate is
\begin{equation}
\widehat r=\frac{\text{distinct novel target hits}/\text{attempts}}
{\text{separately measured seconds}/\text{attempt}}.
\label{eq:useful-rate}
\end{equation}

\begin{figure}[!htbp]
\centering
\includegraphics[width=0.95\linewidth]{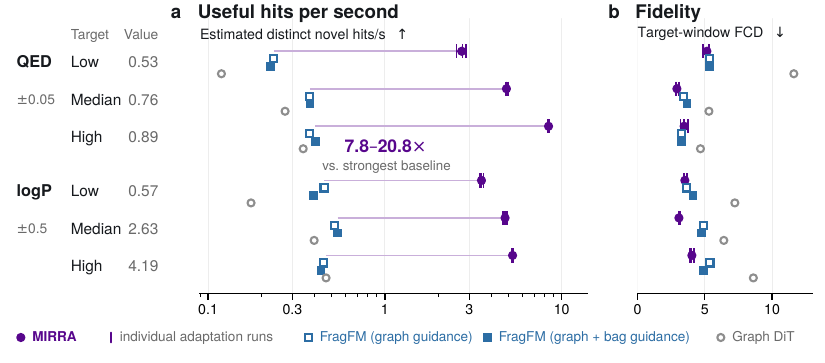}
\caption{\textbf{Faster generation produces more useful property-targeted outputs.}
(a) Estimated distinct novel hits per second; (b) target-window FCD, including
valid misses. Markers average three adapter runs sharing one backbone; ticks show
each run. Rates combine 4,096-attempt yields with three separate 256-attempt
timings; FCD uses 1,000 valid outputs and 1,000 target references. Tolerances
are $\pm0.05$ QED and $\pm0.5$ logP. }
\label{fig:short-conditional}
\end{figure}

\model{} produces an estimated \textbf{7.8--20.8 times} more distinct novel hits
per second than the strongest tested baseline at each target
(Figure~\ref{fig:short-conditional}); the range is 7.8--22.5 across both FragFM
guidance settings and 11.4--24.4 against Graph DiT \citep{graph_dit}. The
advantage does not arise from a higher hit rate: both FragFM settings reach
every target more often per attempt, while \model{} completes 21.3--25.3 times
as many attempts per second as either of them, and 12.3--13.0 times as many as
the fastest baseline at each target (Figure~\ref{fig:figure5-supp}); throughput
more than offsets the lower hit rate. \model{} also has the lowest mean target-window
FCD at five of six targets, with high QED favoring FragFM. These rates average
three adapter runs on one backbone and combine separate yield and timing
populations
(Appendix~\ref{app:conditional-complete}).

\FloatBarrier
\par\addvspace{1.5\baselineskip}
\begin{minipage}{\linewidth}
\section{Discussion}
\model{} makes a large chemical vocabulary practical for fast, high-fidelity
molecular generation. Chemical features let related building blocks share
learning, while identity corrections preserve their differences in molecular
context. Refresh scheduling makes these detailed choices affordable: it
computes full-vocabulary probabilities only at positions where the transition
will use them, preserving the finite-step kernel. Every identity remains
available whenever a position refreshes.

The experiments connect this design to both better distribution matching and
faster generation. Training controls establish the value of chemical alignment,
and generation/assembly interventions show that contextual identity preferences
guide the trajectory itself. In property targeting, higher throughput yields
higher estimated rates of distinct novel hits. This connects generation speed
to the practical yield of molecular search.

Generation uses a finite vocabulary; broader transfer and candidate-specific training remain open.
The central result is that detailed chemical choices can support both fidelity
and speed: chemical knowledge improves their probabilities, and refresh
scheduling removes the cost of predictions that generation does not use.

\label{sec:mainend}
\end{minipage}

\clearpage
\begingroup
\small
\interlinepenalty=10000
\bibliography{references}
\bibliographystyle{plainnat}
\endgroup
\clearpage
\appendix
\setlength{\parskip}{5.5pt}
\setlength{\textfloatsep}{14pt plus 2pt minus 3pt}
\setlength{\intextsep}{12pt plus 2pt minus 2pt}
\renewcommand{\floatpagefraction}{0.7}

\section*{Appendix contents}
\phantomsection\label{app:contents}

\medskip
\begingroup
\renewcommand{\arraystretch}{1.3}
\noindent
\begin{tabular*}{\linewidth}{@{\extracolsep{\fill}}p{0.86\linewidth}r@{}}
\toprule
\textbf{Section and contents} & \textbf{Page}\\
\midrule
\hyperref[app:current-protocol]{\textbf{\ref*{app:current-protocol}\quad Evaluation protocols}} & \pageref{app:current-protocol}\\
\quad Evaluation samples, metrics and timing protocols & \\
\hyperref[app:model-details]{\textbf{\ref*{app:model-details}\quad Model, training and exact scheduling}} & \pageref{app:model-details}\\
\quad Representation, architecture, output algebra, sampler and assembly & \\
\hyperref[app:benchmark-details]{\textbf{\ref*{app:benchmark-details}\quad Benchmark results and comparison protocols}} & \pageref{app:benchmark-details}\\
\quad Step counts, sampling variability, runtime and comparison protocols & \\
\hyperref[app:vocabulary-cost]{\textbf{\ref*{app:vocabulary-cost}\quad Full-vocabulary execution and candidate restriction}} & \pageref{app:vocabulary-cost}\\
\quad Matched cost and fidelity, composition shift and oracle calculation & \\
\hyperref[app:design-details]{\textbf{\ref*{app:design-details}\quad Chemical alignment and identity corrections}} & \pageref{app:design-details}\\
\quad Retrained controls, trajectory/assembly intervention and replications & \\
\hyperref[app:moses-training]{\textbf{\ref*{app:moses-training}\quad Transfer to MOSES}} & \pageref{app:moses-training}\\
\quad Three-seed matched heads, checkpoint sensitivity and novelty & \\
\hyperref[app:conditional-complete]{\textbf{\ref*{app:conditional-complete}\quad Property-conditioned generation}} & \pageref{app:conditional-complete}\\
\quad Adapters, guidance, six-target results, controls and useful rates & \\
\hyperref[app:polymer-training]{\textbf{\ref*{app:polymer-training}\quad Polymer generation}} & \pageref{app:polymer-training}\\
\quad Reserved split, baseline adaptations, metric and internal controls & \\
\bottomrule
\end{tabular*}
\endgroup

\clearpage

\section{Evaluation protocols}
\label{app:current-protocol}
\subsection{ZINC250k benchmark}
ZINC250k is a molecular-generation benchmark derived from the ZINC database
\citep{zinc}.
For results evaluated in this study in Table~\ref{tab:short-published} and
Figures~\ref{fig:short-frontier}--\ref{fig:short-stages}, we score valid molecules
among the first 10,000 attempted outputs against all 24,887 reference molecules,
with reference stereochemistry retained. Failures are not replaced; duplicate
valid molecules remain in FCD and NSPDK. Validity divides valid outputs by
attempts, uniqueness divides unique identities by valid outputs, and novelty
divides unseen unique identities by unique identities against the common
training reference. FCD measures ChemNet feature distributions; NSPDK is the
squared maximum mean discrepancy of labeled local graph features, reported as
MMD$^2\times10^3$ \citep{fcd,nspdk,gdss}.

\model{}250 reports per-metric medians over three sampling draws of one checkpoint. FCD is 0.3665 (range 0.3514--0.3745);
NSPDK is 0.1537 (0.1530--0.1723). \model{}25 and FragFM500 use single draws with FCD 0.3766 and 0.5281, giving reductions of 28.7\% and 30.6\%
for \model{}25 and \model{}250, respectively.

A fixed 10,000-molecule training draw defines the training-data floor:
FCD 0.3361 and NSPDK 0.1096. ZINC informed architecture and correction-rank selection; MOSES and
the reserved polymer reference assess the fixed design.

\subsection{Evaluation protocols for controls and applications}
\paragraph{Independently trained predictors (Table~\ref{tab:short-controls}).}
The matched heads use final exponential moving average (EMA) weights, 250
steps, assembly using training-derived attachment preferences (prior assembly), and the first
25,000 valid outputs from 25,600 attempts, with reference stereochemistry
removed (Appendix~\ref{app:design-details}). Appendix~\ref{app:historical-comparison} reports an additional comparison using the 25,000-valid-output protocol.

\paragraph{Property targeting (Figure~\ref{fig:short-conditional}).}
Three property-adaptation runs share one frozen benchmark backbone; each
baseline regime is one trained model. Each evaluation setting uses 4,096 attempts.
Target-window FCD scores the first 1,000 valid outputs, including misses and
duplicates, against 1,000 held-out molecules inside the requested window.
Distinct novel hits count each valid, in-window, non-training identity once,
ignoring stereochemistry. Each adaptation's yield is divided by its separately
measured time per attempt before rates are averaged. The evaluated property and target settings,
two-seed shuffled-label controls and timing protocol are in
Appendix~\ref{app:conditional-complete}.

\paragraph{Polymers (Table~\ref{tab:short-polymer}).}
The reference contains 9,949 reserved periodic identities; each setting uses
two 4,096-attempt draws of one checkpoint. Structural distances condition on
scorable outputs; validity retains all attempts. MIRRA receives a known linear
chain topology, while the baselines generate complete repeat-unit encodings.
Appendix~\ref{app:polymer-training} specifies the split, periodic
canonicalization, metric and separate validation controls.

\subsection{Timing scopes and uncertainty}
Between-generator timings in Figure~\ref{fig:short-frontier} use batch 256 and
four 4,096-attempt repetitions per primary comparison; the FragFM 250- and 1000-step settings use three repetitions. Figure~\ref{fig:short-vocabulary} instead
uses batch 128 and three 1,024-attempt repetitions per arm. Both include
initialization, generation, transfer, assembly and strict acceptance, excluding
loading, warmup, writing and distribution scoring. Thus 23.32 ms for
\model{}250 in the first study and 43.74 ms in the second refer to different
batching and execution conditions.

Conditional timing uses batch 128 and three 256-attempt repetitions per cell,
including generation, decoding and the common property, identity and hit
checks. Methods within each matched study use the same physical GPU and
balanced execution order, but retain their software versions
(Appendix~\ref{app:protocols}). Sampling variability is measured over draws from fixed checkpoints; training variability is measured over independently trained models.
\FloatBarrier


\clearpage
\section{Model, training and exact scheduling}
\label{app:model-details}
\subsection{Representation and architecture}
A molecule is represented by fragment identities $X=(x_1,\ldots,x_n)$ and a
symmetric binary adjacency matrix $E$. BRICS-derived decompositions expose
typed attachment sites \citep{brics}; identities specify internal atoms and
bonds, while assembly resolves inter-fragment connections. The benchmark vocabulary has 37,821 entries. Matched output-head controls and independent replication models use a separate 38,951-entry vocabulary.

Each identity has a fixed structural and attachment vector $s_v$ combining
normalized Morgan fingerprints and BRICS attachment-type counts.

The graph transformer reads $s_{x_i}$, noisy edges, time and graph structural
features. It has 12 layers, eight attention heads, and node/edge/global widths
384/96/64. A two-layer MLP $g$ maps its contextual node state $z_i$ into the
chemical-feature space, giving vocabulary scores
\begin{equation}
\ell_i(v)=a\,s_v^\top g(z_i)+u_v^\top Dz_i+b_v,
\qquad a=\exp(\tau),\quad p_i=\operatorname{softmax}\ell_i.
\label{eq:head-detail}
\end{equation}
The fixed offset $b_v$ is log training frequency floored at $10^{-12}$.
The correction $UDz_i$ has rank at most 256; $U$ starts at zero. Both branches
score every vocabulary entry. In the 38,951-entry vocabulary, 920 entries share a
feature row with another identity: categorical labels are retained in the state,
but their feature encoding is not injective.

\subsection{Training and compute environment}
\label{app:protocols}
The linear independent-coupling path is
\begin{equation}
q_t(x_t\mid x_1)=t\,\delta_{x_1}(x_t)+(1-t)p_0(x_t).
\label{eq:path-detail}
\end{equation}
Binary edges undergo analogous corruption, sampled on the upper triangle and
mirrored. The objective is mean node cross-entropy plus five times mean edge
cross-entropy, excluding padding and diagonal edges. The source identity
distribution is the empirical fragment-frequency distribution in the training
subset. The binary edge prior uses empirical frequencies over undirected
node pairs, excluding padding and diagonal entries. Graph sizes are drawn
from the empirical distribution of training-graph node counts. These source
distributions remain fixed within each comparison.

The experiments use three model groups: benchmark models for the ZINC
comparisons and the property-adaptation backbone; matched output-head controls
for chemical alignment (Appendix~\ref{app:design-details}); and independent
replication models for the candidate-restriction and identity-correction studies
(Appendices~\ref{app:vocabulary-cost} and \ref{app:head-stages}).

\paragraph{Matched output-head controls.}
The ZINC controls use 213,340 training examples, 11,228 internal-validation
examples, and an 8,192-example checkpoint-selection subset.
They train for 150 epochs at batch 128 (249,900 updates), using AdamW with
learning rate $2\times10^{-4}$, one-epoch warmup, weight decay $10^{-12}$ and
EMA decay 0.9995. Appendix~\ref{app:design-details} distinguishes validation
selection from the common-final comparison.

\paragraph{Benchmark models.}
The benchmark comparison uses the same 208,302 training and 10,926 validation examples for MIRRA and FragFM. Novelty uses 207,114 unique
canonical training identities. MIRRA has 27.2M parameters and trains for 150
epochs (244,050 updates); FragFM uses 100,000 autoencoder and 100,000 flow
updates, with 21.7M flow parameters and a frozen 6.6M-parameter autoencoder.
Data and evaluation are controlled, while architectures, tokenizations and
training recipes differ.

\paragraph{Independent replication models.}
Three models use separate 213,340/11,228 internal
splits, ten-epoch warmup, epoch-wise cosine decay and training times
$t=u^{0.7}$ for uniform $u$. They retain final-epoch EMA. Their initial
log-frequency offset uses the 224,568-example standard training corpus; corruption priors use each training subset.

\paragraph{Compute environment.}
Training and sampling use RTX6000 and H100 GPUs. MIRRA's stack is PyTorch
2.10.0, CUDA 12.8 and RDKit 2025.03.3. Matched unconditional timing uses one RTX6000 per comparison,
four CPU threads, synchronized boundaries and balanced order. Cross-model
throughput includes software differences. Peak GPU memory means
PyTorch allocated tensor memory after warmup, including persistent model
tensors, excluding CPU memory and allocations outside that allocator.

\subsection{What chemical sharing and identity correction express}
\label{app:mechanism-algebra}
Write $\ell=aSh+Uc+b$, with $h=g(z)$, $c=Dz$ and $d=p-e_y$ for target $y$.
Cross-entropy gives
\begin{equation}
\nabla_h L=aS^\top d,\qquad
\nabla_U L=dc^\top,\qquad
\nabla_D L=(U^\top d)z^\top.
\label{eq:head-gradients}
\end{equation}
For a feature-coordinate step $h'=h-\eta\nabla_h L$ holding other factors fixed,
\begin{equation}
\ell'_v-\ell_v=\eta a^2\left(s_v^\top s_y-
                    \sum_w p_w s_v^\top s_w\right).
\label{eq:feature-learning}
\end{equation}
Thus $SS^\top$ couples label errors through chemical similarity. Permuting its
feature rows preserves the spectrum but changes which identities share those
directions. Trained alignment controls test the assignment.

If $s_v=s_w$, the structural head alone fixes their log-odds at $b_v-b_w$.
The correction adds $(u_v-u_w)^\top Dz$, allowing context-dependent odds.
More generally, if $P$ projects onto $\operatorname{col}([\mathbf{1},S])$,
then $(I-P)UDz$ contains score directions the structural branch cannot express,
even allowing a common logit shift. Appendix~\ref{app:correction-probe} tests
their usefulness. A label with no positive optimizer examples still receives
residual gradient $p_vc$ through softmax competition.

The head for the 38,951-entry vocabulary has 10.4M trainable parameters versus 15.1M for a hypothetical
$384\!\to\!384\!\to\!38{,}951$ dense MLP, but requires 2.05 times as many
matrix multiply--accumulates per scored node. Parameter sharing alone does not
establish a runtime advantage; the matched execution experiment does.

\subsection{Sampler and kernel preservation}
Sampling initializes graph sizes, identities and edges from fixed source
laws. On $t_k=k/T$, $k=0,\ldots,T-1$, independently draw node and
upper-triangular edge gates with probability $\alpha_k=1/(T-k)$. For each
graph with any active coordinate, evaluate the backbone on its incoming state.
Evaluate full-vocabulary scores only for active nodes; sample their identities
and the active binary edges, then commit all updates simultaneously and mirror
the edges. An active node can resample its current identity. Graphs without
active coordinates require no forward pass.

This realizes exactly
\begin{equation}
K_{i,k}(v\mid X,E)=(1-\alpha_k)\mathbf{1}[v=x_i]
             +\alpha_k p_{\theta,i}(v\mid X,E,t_k)
\label{eq:refresh-kernel-detail}
\end{equation}
in real arithmetic: a zero gate retains the identity regardless of its unused
prediction; a one gate draws from the complete posterior. Gate independence
permits the computation order to change. Graphs can also be batched at their
next active times because inference has no interaction between batch members.

For $n$ nodes and $m=n(n-1)/2$ undirected edges, the expected numbers of node
projections and graph evaluations are respectively
\begin{equation}
nH_T=n\sum_{j=1}^{T}\frac1j,
\qquad \sum_{k=0}^{T-1}\left[1-(1-\alpha_k)^{n+m}\right].
\end{equation}
The first replaces $nT$ dense projections. The backbone and edge head remain
dense within active graphs; batching and assembly still cost time.
Rebatching changes floating-point arithmetic, so kernel equality does not imply
bitwise-identical trajectories. Floating-point differences occasionally change categorical draws near CDF boundaries. Their quality is measured
separately in Appendix~\ref{app:vocabulary-cost}.

\subsection{Measured refresh counts}
\label{app:refresh-counts}
\begin{figure}[!htbp]
\centering
\includegraphics[width=\linewidth]{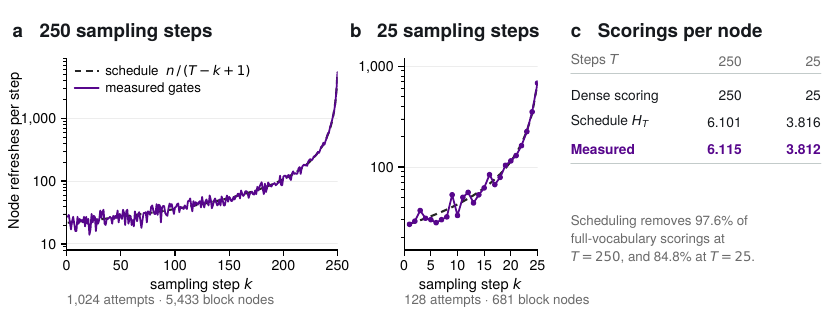}
\caption{\textbf{Refresh scheduling reduces expected full-vocabulary evaluations per node from $T$ to $H_T$.}
(a) At $T=250$, node refreshes per step against the schedule's expectation
$n/(T-k+1)$ for $n$ block nodes and one-based step index $k=1,\ldots,T$,
on a logarithmic axis. Early steps refresh few
positions and the final step refreshes every one.
(b) The same at $T=25$.
(c) Full-vocabulary evaluations per node. Dense scoring evaluates the vocabulary at
every step; the schedule predicts the harmonic number $H_T$; the measured counts are total refreshes divided by the number of block nodes. Sampling uses 1,024 attempts at $T=250$ and 128 at $T=25$ from the benchmark checkpoint.}
\label{fig:app-refresh}
\end{figure}

At both 250 and 25 steps, measured mean refresh counts per node agree with the harmonic prediction $H_T$ (Figure~\ref{fig:app-refresh}).

\FloatBarrier

\subsection{Assembly decoder}
\label{app:assembly}
At $k=T-1$, every node refreshes. The complete node and edge probabilities
evaluated on the final incoming graph are retained for assembly. The decoder
receives these probabilities, the sampled terminal identities and the graph
size. Terminal edge probabilities supply connection scores for the tree search.

\paragraph{Attachment-count feasibility.}
For $n>1$, the current identities must expose at least one attachment site per
block and $2(n-1)$ sites in total
(Equation~\ref{eq:assembly-constraints}). When these requirements are unmet,
the decoder attempts at most $n$ identity adjustments. Positions are initially
ordered by increasing maximum node probability. Blocks with no attachment sites
are considered first; otherwise, the decoder seeks a position whose replacement
can balance the total count. It selects the highest-probability identity in the
chosen attachment-count group and moves the edited position to the end of the
order. The desired count is restricted to $1,\ldots,n-1$. If the count
requirements remain unmet, the attempt fails before tree search. A single-block
output requires no attachment sites and a valid molecule.

\paragraph{Tree search and site pairing.}
Connection scores are the terminal edge log-odds,
$w_{ij}=\log p^E_{ij}(1)-\log p^E_{ij}(0)$, with probabilities floored at
$10^{-9}$ for this calculation. Candidate trees have degrees equal to the
attachment counts. A greedy construction first considers edges in decreasing
score order, subject to degree limits and cycle avoidance. When this produces
a spanning tree, it is placed first in the candidate list. Additional
degree-exact trees are obtained from 96 randomized Pr\"ufer-sequence draws,
deduplicated and ordered by summed edge score. They are added to the list up
to a total of 48 candidates. Each tree is checked in order, stopping at the
first successful assembly.

For each candidate tree, the decoder enumerates at most 1,000 compatible
attachment pairings. The primary policy scores these pairings using
training-derived attachment preferences and tests the highest-scoring pairing
for valid molecular assembly. The canonical policy instead considers the
valid molecules produced by the enumerated pairings and selects the one with
the lexicographically smallest canonical SMILES. This policy also defines the
alternate assembly arm used in the controls.

\paragraph{Identity repair and failed attempts.}
If the attachment counts are feasible but tree assembly is unsuccessful, the
decoder tries single-identity replacements at up to three positions with the
lowest maximum node probabilities. At each position, it considers up to eight
highest-probability identities with the same attachment count, skipping the
current identity. Each candidate is tried separately and the first successful
replacement is accepted. These repair attempts use canonical assembly,
including when the initial search used the attachment prior.
Unsuccessful attempts remain failures and are not replaced.
Appendix~\ref{app:head-stages} separates identity correction during the
trajectory from its use in terminal assembly scores.

\section{Benchmark results and comparison protocols}
\label{app:benchmark-details}
\subsection{Operating points and sampling variability}
Table~\ref{tab:app-operating} reports the operating points in Figure~\ref{fig:short-frontier}. MIRRA has lower FCD and NSPDK at both shared
step counts, 250 and 1000; the separate matched timings establish its practical
advantage at 25 and 250 steps. FragFM retains higher novelty. All rows use the 10,000-attempt protocol of Appendix~\ref{app:current-protocol}.
\begin{table}[!htbp]
\caption{\textbf{ZINC250k results across sampling step counts.} Both generators use the same training data. Bold marks the best value in each column, using unrounded values.}
\label{tab:app-operating}
\centering
\medskip
\begingroup
\small
\setlength{\tabcolsep}{2pt}
\begin{tabular*}{\linewidth}{@{\extracolsep{\fill}}lrrrrrrr@{}}
\toprule
Setting & Valid $n$ & FCD $\downarrow$ & NSPDK $\downarrow$ & Valid & Unique & Novel & ms/attempt \\
& & & $(\times10^3)$ & (\%) & (\%) & (\%) & $\downarrow$ \\
\midrule
\multicolumn{8}{@{}l}{\emph{Common training data, one evaluation protocol}}\\[1pt]
\model{}, 25 steps & 9,927 & 0.3766 & 0.1795 & 99.27 & \textbf{99.98} & 97.69 & \textbf{15.7} \\
\model{}, 250 steps & 9,980--9,985 & \textbf{0.3665} & \textbf{0.1537} & 99.81 & 99.97 & 97.61 & 23.3 \\
\model{}, 1000 steps & \textbf{9,988} & 0.3824 & 0.1613 & \textbf{99.88} & 99.97 & 97.61 & -- \\
\addlinespace[2pt]
FragFM, 250 steps & 9,878 & 0.5296 & 0.3782 & 98.78 & 99.97 & 99.22 & 81.5 \\
FragFM, 500 steps & 9,914 & 0.5281 & 0.3824 & 99.14 & 99.98 & 99.23 & 132.4 \\
FragFM, 1000 steps & 9,890 & 0.5167 & 0.4010 & 98.90 & 99.97 & \textbf{99.27} & 270.6 \\
\bottomrule
\end{tabular*}
\par\smallskip
\begin{minipage}{\linewidth}\footnotesize
Every row scores the valid subset of the first 10,000 attempts against all 24,887 references, with failures not replaced. \model{} at 250 steps is a per-metric median of three sampling draws; all other rows are single draws. Times are medians of four 4,096-attempt repetitions, except the FragFM 250- and 1000-step settings (three repetitions); their scope and compute setup are defined in Appendix~\ref{app:current-protocol}. Quality and timing use separate populations. FragFM500 timing is from the 250-versus-500 study; the separately matched 25-versus-500 study appears in Table~\ref{tab:app-timing}. At the shared 250- and 1000-step settings \model{} has lower FCD and NSPDK; no matched 1000-step MIRRA timing is reported. FragFM has higher novelty at all reported settings.
\end{minipage}
\endgroup
\end{table}

Eight sampling draws of the benchmark checkpoint at 250 steps give mean FCD
0.3659 and sample SD 0.0068 (range 0.3514--0.3745), compared with FragFM500's
single-draw 0.5281. Mean NSPDK is $0.1610\pm0.0073$; mean validity,
uniqueness and novelty are 99.82\%, 99.94\% and 97.54\%. The first three draws supply the per-metric medians. The spread describes sampling variability from one checkpoint. The checkpoint's
step sweep is not monotone: FCD improves from 25 to 250 steps, then rises
slightly at 1000.

\subsection{Matched generation time and memory}
Table~\ref{tab:app-timing} reports matched generation times across sampling step counts. Timers include all attempts and
one assembly policy, with the scope in Appendix~\ref{app:current-protocol}.
The matched 25-versus-500 experiment gives median times 15.69 and 134.53 ms,
a factor of 8.58. The separate 250-versus-500 experiment gives 23.32 and
132.43 ms, a factor of 5.68. Quality populations are separate from timing runs.

In the 250-versus-500 study, median peak PyTorch allocation is
\TimingCurrentAllocatedGiB{} versus \TimingFragFMAllocatedGiB{} GiB,
\TimingMemoryReduction{}\% lower for MIRRA. This counter includes persistent
model tensors but excludes loading and warmup peaks, CPU memory and allocations
outside PyTorch.
\begin{table}[!htbp]
\caption{\textbf{Matched generation times.} Individual timing repetitions and their medians.}
\label{tab:app-timing}
\centering
\medskip
\begingroup
\small
\setlength{\tabcolsep}{3pt}
\begin{tabular*}{\linewidth}{@{\extracolsep{\fill}}lrrrrrr@{}}
\toprule
Arm & Attempts & Rep 1 & Rep 2 & Rep 3 & Rep 4 & Median \\
& per rep & (ms) & (ms) & (ms) & (ms) & (ms) \\
\midrule
\multicolumn{7}{@{}l}{\emph{Matched 25 vs.\ 500 steps}}\\[1pt]
\model{}, 25 steps & 4,096 & 26.31 & 15.77 & 15.61 & 15.04 & \textbf{15.69} \\
FragFM, 500 steps & 4,096 & 267.43 & 144.19 & 122.19 & 124.86 & \textbf{134.53} \\
\addlinespace[2pt]
\multicolumn{7}{@{}l}{\emph{Matched 250 vs.\ 500 steps}}\\[1pt]
\model{}, 250 steps & 4,096 & 23.40 & 22.90 & 23.24 & 24.48 & \textbf{23.32} \\
FragFM, 500 steps & 4,096 & 132.20 & 131.52 & 132.67 & 136.39 & \textbf{132.43} \\
\addlinespace[2pt]
\multicolumn{7}{@{}l}{\emph{FragFM, 250 steps}}\\[1pt]
FragFM, 250 steps & 4,096 & 81.54 & 79.19 & 99.77 & -- & \textbf{81.54} \\
\addlinespace[2pt]
\multicolumn{7}{@{}l}{\emph{FragFM, 1000 steps}}\\[1pt]
FragFM, 1000 steps & 4,096 & 323.44 & 264.70 & 270.58 & -- & \textbf{270.58} \\
\bottomrule
\end{tabular*}
\par\smallskip
\begin{minipage}{\linewidth}\footnotesize
Milliseconds per attempted molecule, covering generation, transfer, single-policy assembly and strict acceptance after warmup, and excluding loading, warmup, output writing and distribution scoring. Within each study both arms share the same GPU and batch size 256, in balanced order. \model{} and FragFM retain different software versions.
\end{minipage}
\endgroup
\end{table}

\subsection{Metric conventions}
\label{app:published}
FCD is sensitive to population size and molecular representation. In a separate
preprocessing sensitivity analysis, a stereo-stripped training population against the reference
scores 0.3361, compared with 0.1881 for a stereo-retaining population. Passing
molecules through a simplified element-only graph reconstruction raises the
training score to 1.1183, close to the published 1.13 anchor; the same
transformation gives 1.1118--1.1332 for the independent replication models. This transformation changes approximately one third of training molecules, illustrating the sensitivity of FCD to molecular representation. NSPDK's element/bond labels give a much more
stable training anchor, $0.1096\times10^{-3}$, near the published
$0.10\times10^{-3}$.

Our NSPDK implementation follows GDSS: atom-symbol and integer bond-order
labels without kekulization; EDeN complexity four, discrete features, 16-bit
hashing, inner/global normalization and a fixed Python hash seed. The biased
linear-kernel MMD$^2$, including diagonal terms, is the squared distance
between mean feature vectors. These graph features complement the ChemNet features used by FCD.

\subsection{25,000-valid-output and released-checkpoint comparisons}
\label{app:historical-comparison}
\label{app:samplingsteps}
\label{app:released-timing}
The comparison using the benchmark models uses 25,600 attempts, FCD on the first 25,000 valid outputs and the 24,887-entry stereo-stripped reference.
MIRRA250 and locally trained FragFM500 score 0.1584 and 0.3744, a 57.7\%
reduction; NSPDK is 0.1131 and 0.3272 ($\times10^3$).
Validity favors MIRRA; uniqueness and novelty favor FragFM. All 16 comparisons across assembly rules, step counts and valid-output sample sizes favor MIRRA on both distances.
For charge-projected FCD, the same transformation is applied to generated
and reference molecules: negatively charged atoms are converted to their
protonated neutral form, while positive charges are retained. This gives FCD
0.1585 versus 0.2875, retaining a 44.9\% advantage. The 25-step setting gives FCD 0.1792 with prior assembly.

The authors' released FragFM checkpoint is a separate comparator trained on a
split containing 93.9\% of the evaluation-reference rows. Table~\ref{tab:app-released-frontier}
reports its tested operating points and the MIRRA benchmark checkpoint.
Each time is the median of four 1,024-attempt repetitions at batch 256 on the same GPU, in rotated order after 256 warmup attempts. Both methods include
initialization, sampling, transfer and decoding; loading, writing and scoring
are excluded. Their decoders and software versions remain different.
\begin{table}[!htbp]
\caption{\textbf{Fidelity and measured cost against the released FragFM
checkpoint.} Both measured \model{} settings reach lower FCD and lower median
generation time than the three tested released settings.}
\label{tab:app-released-frontier}
\centering
\medskip
\begingroup
\small
\setlength{\tabcolsep}{3pt}
\begin{tabular*}{\linewidth}{@{\extracolsep{\fill}}lrrrr@{}}
\toprule
Model & Steps & Median (s) & FCD $\downarrow$ & Projected FCD $\downarrow$ \\
\midrule
\model{} & 25 & \ReleasedMirraFastSeconds{} & 0.1792 & 0.1814 \\
\model{} & 250 & \ReleasedMirraMainSeconds{} & 0.1584 & 0.1585 \\
Released FragFM & 100 & \ReleasedFragFastSeconds{} & 0.4220 & 0.3071 \\
Released FragFM & 250 & \ReleasedFragBestSeconds{} & 0.3261 & 0.2228 \\
Released FragFM & 500 & \ReleasedFragLongSeconds{} & 0.3356 & 0.2255 \\
\bottomrule
\end{tabular*}
\par\smallskip
\begin{minipage}{\linewidth}\footnotesize
Seconds per 1,024 raw attempts including assembly, as the median of four
repetitions; the two FCD columns retain the molecular scoring conventions
before and after the charge projection defined above. Each FCD value is computed from one generated sample; times are medians of four separate timing runs. The released
and benchmark checkpoints use different training splits.
\end{minipage}
\endgroup
\end{table}

\subsection{Consecutive molecular samples}
\begin{figure}[!htbp]
\centering
\includegraphics[width=\linewidth,height=0.72\textheight,keepaspectratio]{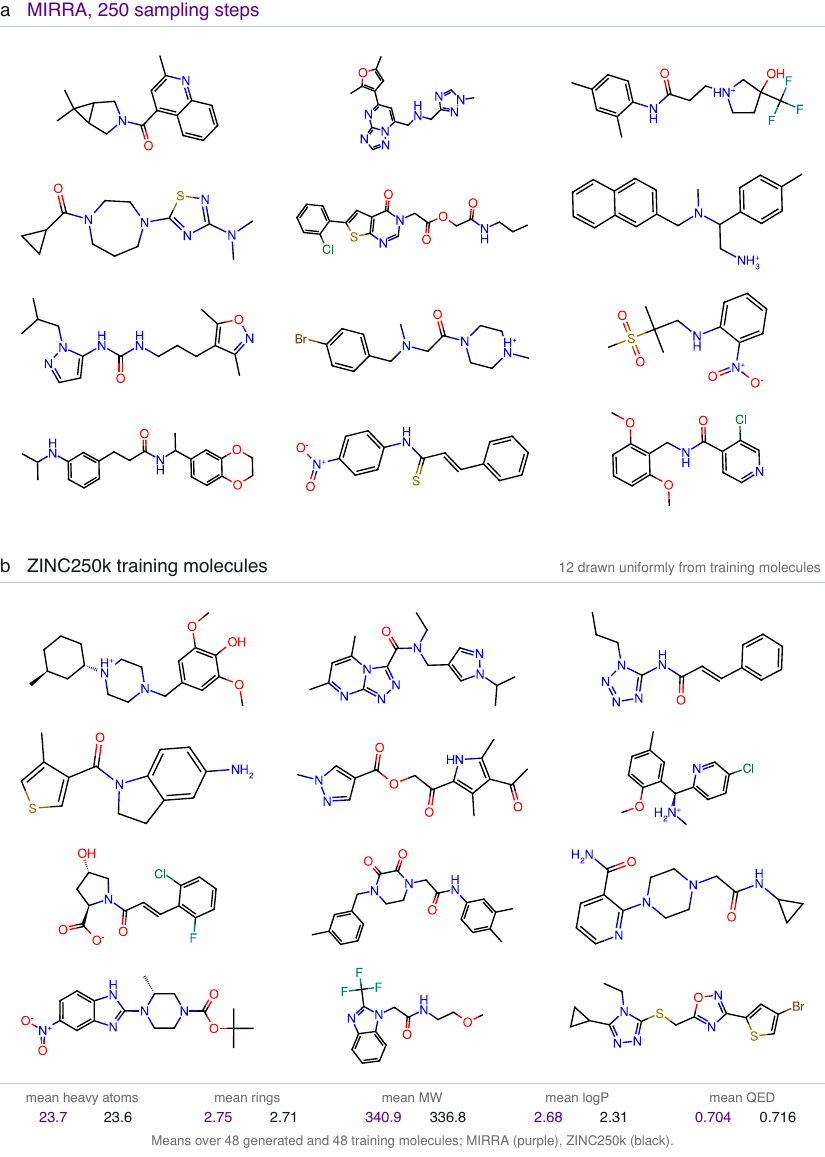}
\caption{\textbf{Consecutive \model{} outputs beside training molecules.}
(a) The first 12 outputs of one 250-step run of the benchmark checkpoint,
shown in generation order without filtering, deduplication or manual selection.
All 12 are chemically valid and connected.
(b) 12 molecules drawn uniformly at random, without replacement, from the training molecules.
The strip beneath reports descriptor means over the first 48 outputs of the same run and 48 uniformly drawn training
molecules. }
\label{fig:app-samples}
\end{figure}

Figure~\ref{fig:app-samples} compares consecutive generated molecules with training molecules and summarizes their structural descriptors.

\section{Full-vocabulary execution and candidate restriction}
\label{app:vocabulary-cost}
\subsection{Dense, scheduled and restricted scoring}
This comparison uses the 37,821-entry benchmark checkpoint, 250 steps,
and the 10,000-attempt quality protocol. The three arms share weights, source
laws and assembly: dense full-vocabulary evaluation; scheduled full-vocabulary
evaluation; and scheduled candidate-only evaluation. The last draws 1,536
distinct identities without replacement from the training prior per scheduled
round and adds current graph identities, so actual set sizes can exceed 1,536.
It gathers and projects only candidate columns at nonterminal updates and
renormalizes over that set. It uses neither retraining nor a proposal correction.
Its sampled terminal state follows the restricted transition, while assembly
receives complete terminal posteriors in every arm.

Three quality draws per arm give the fidelity ranges in
Table~\ref{tab:app-vocabulary-full}. Three separate 1,024-attempt timing runs
rotate arm order on the same GPU at batch 128 after warmup, including candidate
construction, transfer and assembly. Scheduling reduces median time from
131.56 to 43.74 ms per attempt at essentially unchanged FCD. Candidate scoring
uses less peak memory (0.556 versus 0.853 GiB), but takes 46.18 ms and raises
FCD to 0.9469. Its higher novelty accompanies a shifted distribution.
\begin{table}[!htbp]
\caption{\textbf{Exact refresh scheduling preserves full-support quality; candidate restriction changes it.} One benchmark checkpoint at 250 steps.}
\label{tab:app-vocabulary-full}
\centering
\medskip
\begingroup
\small
\setlength{\tabcolsep}{3pt}
\begin{tabular*}{\linewidth}{@{\extracolsep{\fill}}lrrrrr@{}}
\toprule
Setting & FCD $\downarrow$ & NSPDK $\downarrow$ & Valid (\%) & Unique (\%) & Novel (\%) \\
\midrule
Dense, full vocabulary & 0.3680 & 0.1543 & 99.81 & 99.97 & 97.62 \\
Refresh, full vocabulary & 0.3665 & 0.1537 & 99.81 & 99.97 & 97.61 \\
Refresh, sampled subset & 0.9469 & 0.8780 & 99.76 & 99.67 & 98.14 \\
\midrule
\multicolumn{6}{@{}l}{\emph{End-to-end timing and peak allocated memory}}\\[2pt]
Setting & \multicolumn{2}{c}{ms/attempt, median [range]} & \multicolumn{3}{c}{Peak allocation (GiB)} \\
Dense, full vocabulary & \multicolumn{2}{c}{131.56 [125.52, 191.77]} & \multicolumn{3}{c}{1.364} \\
Refresh, full vocabulary & \multicolumn{2}{c}{43.74 [42.64, 52.43]} & \multicolumn{3}{c}{0.853} \\
Refresh, sampled subset & \multicolumn{2}{c}{46.18 [44.23, 48.70]} & \multicolumn{3}{c}{0.556} \\
\bottomrule
\end{tabular*}
\par\smallskip
\begin{minipage}{\linewidth}\footnotesize
Quality entries are per-metric medians of three 10,000-attempt draws against 24,887 references; NSPDK is MMD$^2\times10^3$. FCD ranges are 0.3510--0.3741, 0.3514--0.3745 and 0.9057--0.9690 in row order. Timing uses three separate 1,024-attempt repetitions at batch 128, including assembly and acceptance; peak allocation is the maximum across repetitions. These timings have a different batch and execution scope from the between-generator study.
\end{minipage}
\endgroup
\end{table}

\subsection{Composition shifts occur before assembly}
\label{app:support}
A separate replication on three independent replication models with 38,951 entries uses
one 25,000-valid population per arm and model, with full terminal assembly
scores. Restriction raises mean FCD from 0.1500 to 0.6925, lowers mean
heavy-atom count from 23.07 to 21.77 and molecular weight from 331.1 to
311.7, while raw validity remains above 99.7\%. Ten of eleven descriptor
distances worsen; logP is the exception.

Paired comparisons use 512 attempts per arm and model. Among pairs for which both outputs are valid, the restricted-minus-full heavy-atom difference averages $-1.145$ before assembly and $-1.103$ in the final molecule. Attachment-count adjustment and assembly contribute $-0.016$ and $+0.058$, respectively. Across all attempts, the terminal difference is $-1.173$. The size shift therefore precedes assembly.

At a fixed incoming state and realized candidate set $S$, write $m=p(S)$ and
$p^S=p(\cdot\mid S)$. Then
\begin{equation}
D_{\mathrm{TV}}(p,p^S)=1-m,\qquad
D_{\mathrm{TV}}(K,K^S)=\alpha_k(1-m).
\end{equation}
For a fragment property $h$, with posterior means $\mu_S$ and $\mu_{S^c}$,
\begin{equation}
\mathbb{E}_{p^S}h-\mathbb{E}_{p}h=(1-m)(\mu_S-\mu_{S^c}).
\label{eq:support-drift}
\end{equation}
Across 4,114 recorded node refreshes per model, mean omitted
posterior mass is 4.75--5.16\%, and expected heavy-atom count shifts by
$-0.195$ to $-0.215$ per refresh. These measurements share graphs, candidate sets and trajectories.

\subsection{An oracle example: smaller steps do not remove proposal bias}

\label{app:oracle-support}
Consider one coordinate with $V\geq2$ identities, strictly positive source and
target both $p$, and the linear conditional path
$q_t(x\mid y)=(1-t)p(x)+t\mathbf{1}[x=y]$. Its marginal is $p$ for every
$t$, and its exact Bayes endpoint posterior is
$p_{1\mid t}(v\mid x)=(1-t)p(v)+t\mathbf{1}[v=x]$.
At each refresh, independently draw one candidate from a fixed strictly
positive distribution $\nu$, add the current identity, and renormalize this
posterior over the resulting set.

\textbf{Proposition.} All candidate-averaged transition kernels share the
invariant distribution
\[
\pi_i=p_i\nu_i/\sum_jp_j\nu_j.
\]
Their time-one endpoint
equals $p$ on every uniform grid, and in the continuous-time limit, if and only
if $\nu$ is uniform. Full-support refreshes preserve $p$ on every grid.

\textit{Proof.} A transition from $i$ to $j\ne i$ requires drawing candidate
$j$. The retained posterior mass is $d_{ij}(t)=t+(1-t)(p_i+p_j)$, giving
the averaged kernel $K_k=I+G_{t_k}/T$, where
\begin{equation}
G_t(i,j)=\frac{\nu_jp_j}{d_{ij}(t)}\quad(i\ne j),\qquad
G_t(i,i)=-\sum_{j\ne i}G_t(i,j).
\label{eq:oracle-candidate-generator}
\end{equation}
Symmetry of $d_{ij}$ gives detailed balance:
$\pi_iG_t(i,j)=\pi_jG_t(j,i)$ for every $t$. Thus $\pi K_k=\pi$.
All entries of $K_k$ are positive, including the diagonal, since drawing
the current identity forces a stay. The strict log-sum inequality gives
$D_{\mathrm{KL}}(\mu K_k\Vert\pi)<D_{\mathrm{KL}}(\mu\Vert\pi)$ whenever
$\mu\ne\pi$. If $\nu$ is nonuniform, $p\ne\pi$, so a trajectory starting at
$p$ cannot return to $p$ after any positive number of steps. If $\nu$ is
uniform, $\pi=p$ and every kernel preserves it.

The rates in Equation~\ref{eq:oracle-candidate-generator} are bounded and smooth
on $[0,1]$, so grid refinement converges to $\dot\mu=\mu G_t$. With
$r_i=\mu_i/\pi_i$, its relative entropy satisfies
\begin{equation}
\frac{d}{dt}D_{\mathrm{KL}}(\mu\Vert\pi)
=-\frac12\sum_{i\ne j}\pi_iG_t(i,j)(r_i-r_j)(\log r_i-\log r_j),
\end{equation}
which is strictly negative away from $\pi$. Hence the time-one limit also
differs from $p$ for nonuniform $\nu$. Finally,
$p\,p_{1\mid t}=p$ implies preservation under full-support refreshes.
This proves the proposition.

For a prior proposal $\nu=p$, the common invariant is proportional to $p_i^2$;
it is generally \emph{not} the finite-time endpoint.

\paragraph{Two-identity closed form.}
Set $p=\nu=(a,1-a)$. If the candidate equals the current identity no change
is possible; otherwise both identities are available. Since
$\alpha_k(1-t_k)=1/T$, averaging over the candidate gives the same kernel at
every step:
\begin{equation}
K^S=\begin{pmatrix}
1-(1-a)^2/T & (1-a)^2/T\\
a^2/T & 1-a^2/T
\end{pmatrix}.
\end{equation}
Writing $c=a^2+(1-a)^2$ and $\pi=a^2/c$, the probability of the first identity
after all $T$ steps is
\begin{equation}
\Pr(X_T=0)=\pi+(a-\pi)(1-c/T)^T
\;\longrightarrow\;\pi+(a-\pi)e^{-c}.
\end{equation}
For $a=0.8$, the restricted endpoint tends to $(0.870,0.130)$, a total variation of 0.070 from the target $(0.8,0.2)$. Uniform candidate proposals preserve the target, and reversing the proposal weights reverses the direction of the mass shift. Restriction therefore need not bias every proposal, but time refinement alone does not generally remove its bias.

\FloatBarrier

\section{Chemical alignment and identity corrections}
\label{app:design-details}
\subsection{Retrained output-head controls}
Table~\ref{tab:short-controls} fixes input features, backbone, full categorical
support, training budget and assembly, while changing the output predictor.
All models use the 38,951-entry vocabulary and matched recipe in
Appendix~\ref{app:protocols}; FCD and NSPDK use the first
25,000 valid outputs and the stereo-stripped reference. These experiments test
learning, separately from the fixed-weight interventions below.

The broad shuffle replaces output row $s_v$ by $s_{\pi(v)}$ within positive-
and zero-prior strata. It preserves the feature multiset, residual correction,
fixed priors and all 27.5M parameters, while breaking feature-to-label and
feature-to-frequency associations. Correct alignment lowers primary FCD by
10.63--15.74\% across three independently trained pairs
(Table~\ref{tab:app-alignment-replication}). The narrower shuffle preserves
exact attachment signatures and log2 frequency bins, changing fingerprints for identities accounting for 37.32\% of training occurrences; exact within-bin frequencies are not matched.
Its FCD 0.1655 versus 0.1561 aligned gives a 5.72\% reduction at 250 steps,
with an 8.25\% reduction at 1000. Both distances favor alignment in all eight comparisons across assembly rules, step counts and valid-output sample sizes.

\begin{table}[!htbp]
\caption{\textbf{Chemical alignment improves fidelity in three independently
trained pairs.} Primary 250-step, prior-assembly comparison.}
\label{tab:app-alignment-replication}
\centering\small
\begin{tabular*}{\linewidth}{@{\extracolsep{\fill}}lrrr@{}}
\toprule
Training pair & Aligned FCD & Shuffled FCD & Reduction (\%)\\
\midrule
A & 0.1561 & 0.1757 & 11.19\\
B & 0.1543 & 0.1727 & 10.63\\
C & 0.1551 & 0.1841 & 15.74\\
\bottomrule
\end{tabular*}
\par\smallskip
\begin{minipage}{\linewidth}\footnotesize
Each pair uses one generation draw per arm. Mean NSPDK$\times10^3$ is
$0.1117\pm0.0073$ aligned versus $0.1368\pm0.0156$ shuffled at 250 steps,
and $0.1121\pm0.0079$ versus $0.1278\pm0.0110$ at 1000 (sample SD over
models). Both distances favor alignment in all three primary pairs. Execution
is matched within each pair; pooled spread also includes hardware and execution
differences across pairs.
\end{minipage}
\end{table}

\paragraph{Conventional head and checkpoint selection.}
A $384\!\to\!264\!\to\!38{,}951$ ReLU MLP with biases and the same fixed
log-frequency offset matches the parameter count within 0.02\%. Structural
inputs, backbone initialization, training streams and budget are held fixed;
output architecture, initialization and score rank change. Minimum EMA
validation loss selects update 40,000 for this head and 249,900 for MIRRA.
The selected FCDs are 0.4001 versus 0.1561; at the common final update they
are 0.1793 versus 0.1561. At 1000 steps, final FCD is 0.1695 versus 0.1500.
The main text uses the common-final comparison. An additional weight-decay-0.1
control improves the conventional selected FCD to 0.2625, but worsens its
final FCD to 0.2077. MIRRA stays ahead under both policies. The conventional
head retains slightly higher novelty (98.05\% versus 97.81\% at final EMA,
250 steps).

\paragraph{Reference overlap and representation checks.}
Removing 285 exact training identities from the matched-control
reference leaves 24,602 molecules and preserves every comparison's direction:
primary FCD is 0.1587 aligned versus 0.1776 shuffled. Separately, four fragmentation schemes on the same training data
cross implicit/explicit-hydrogen cut selection with all/single-bond BRICS cuts.
Their 250-step prior-assembly FCDs span 0.1537--0.1640. The explicit-H,
single-bond variant reproduces FragFM's mean training fragment count (6.12)
and scores 0.1640 versus 0.3744 for FragFM500. Vocabulary, capacity
and graph size change together; this checks robustness to representation
choices rather than isolating fragment count.

\subsection{Where identity corrections affect generation}
\label{app:head-stages}
The benchmark intervention independently enables the learned identity
correction during the generation trajectory and at final assembly. Within a
trajectory, the two assembly conditions share terminal states, edge posteriors
and paired assembly randomness; only terminal node scores change. Three
10,000-attempt draws per condition use the benchmark reference and are summarized
in Table~\ref{tab:app-stages}.

Holding assembly correction on, enabling trajectory correction reduces FCD by
95.1--95.4\% across the three primary draws. Assembly-only correction raises
validity by about eight percentage points without recovering FCD. With full
trajectory correction, assembly effects on FCD are small and mixed. Repeating
the experiment on a second GPU type with additional sampling draws retains the dominant
trajectory effect. These comparisons measure reliance of fixed coadapted
weights; the independently trained heads above address the separate learning
question.
\begin{table}[!htbp]
\caption{\textbf{Corrections during generation dominate the fidelity improvement on both devices.} The assembly effect is smaller and metric-dependent.}
\label{tab:app-stages}
\centering
\medskip
\begingroup
\small
\setlength{\tabcolsep}{3pt}
\begin{tabular*}{\linewidth}{@{\extracolsep{\fill}}lrrrrr@{}}
\toprule
Correction active at & FCD, median [range] $\downarrow$ & NSPDK $\downarrow$ & Valid & Unique & Novel \\
& & $(\times10^3)$ & (\%) & (\%) & (\%) \\
\midrule
\multicolumn{6}{@{}l}{\emph{RTX6000}}\\[2pt]
Neither & 7.4210 [7.4134, 7.4297] & 16.8574 & 80.03 & 96.87 & 98.59 \\
Assembly only & 7.6384 [7.5242, 7.6598] & 13.9544 & 88.10 & 89.52 & 98.54 \\
Generation only & 0.3675 [0.3538, 0.3726] & 0.1600 & 99.68 & 99.97 & 97.65 \\
Generation + assembly & 0.3665 [0.3514, 0.3745] & 0.1537 & 99.81 & 99.97 & 97.61 \\
\midrule
\multicolumn{6}{@{}l}{\emph{H100}}\\[2pt]
Neither & 7.4976 [7.4637, 7.6855] & 17.2978 & 80.51 & 96.56 & 98.64 \\
Assembly only & 7.4563 [7.4522, 7.7772] & 13.5305 & 88.03 & 89.59 & 98.55 \\
Generation only & 0.3696 [0.3659, 0.3726] & 0.1592 & 99.70 & 99.95 & 97.57 \\
Generation + assembly & 0.3734 [0.3713, 0.3735] & 0.1614 & 99.81 & 99.95 & 97.48 \\
\bottomrule
\end{tabular*}
\par\smallskip
\begin{minipage}{\linewidth}\footnotesize
Each device uses three 10,000-attempt draws per condition from the same checkpoint at 250 steps. All other quality entries are per-metric medians; FCD uses the common 24,887-entry reference. Assembly-only versus neither, and both versus generation-only, need not have the same sign for every metric.
\end{minipage}
\endgroup
\end{table}

A separate replication uses three independent replication models, 4,096
attempts per cell and the first 3,000 valid molecules against an 11,228-molecule
internal reference. With full assembly scoring fixed, enabling trajectory corrections reduces FCD by 92.1--92.5\% across the three models. Using the first 1,000 valid outputs and the alternate assembly rule, every trajectory comparison also favors correction.

\subsection{What the correction adds to the output probabilities}
\label{app:correction-probe}
At a fixed contextual state, write the base logit as
$b(v)=a s_v^\top g(z)+b_v$ and its learned correction as $r(v)=u_v^\top Dz$.
Let $B=\operatorname{softmax}(b)$ and
$F=\operatorname{softmax}(b+r)$. The attachment groups partition the
vocabulary by the multiset of typed attachment sites; the 38,951-entry ZINC
vocabulary has 452 such groups. These groups need not have identical
structural feature vectors. Define
\begin{equation}
A_g=\log\sum_{v:G(v)=g}e^{b(v)},\qquad
C_g=\log\sum_{v:G(v)=g}e^{b(v)+r(v)},\qquad
\Delta_g=C_g-A_g.
\label{eq:group-correction}
\end{equation}
Then $\Delta_g=\log\mathbb{E}_{v\sim B(\cdot\mid g)}e^{r(v)}$ and
\begin{equation}
F(g)=\frac{B(g)e^{\Delta_g}}{\sum_h B(h)e^{\Delta_h}},\qquad
F(v\mid g)=B(v\mid g)e^{r(v)-\Delta_g}.
\label{eq:correction-factors}
\end{equation}
These identities follow by summing the exponentiated logits within each
group. A single identity correction therefore induces both a group
reweighting and a conditional identity reweighting. The group-only
intervention uses logits $b(v)+\Delta_{G(v)}$; identity-only uses
$b(v)+r(v)-\Delta_{G(v)}$. Their normalized probabilities are exactly
$F(G(v))B(v\mid G(v))$ and $B(G(v))F(v\mid G(v))$, respectively, in real
arithmetic. A correction constant within each group changes only group
allocation. A correction with every $\Delta_g=0$ changes only conditional
identity probabilities. In general, both change.

Group log-sums are computed in float64 and returned in model precision; the
four probability factors match Equation~\ref{eq:correction-factors} to
numerical precision. Table~\ref{tab:head} measures the generation effects of these interventions.

The group-only and identity-only posteriors replace the predictor throughout
inference, including generation and terminal assembly.

Table~\ref{tab:head} compares all four arms using the first 3,000 valid outputs
from each condition's 4,096 attempts, the common evaluation size supported by
all four conditions, without replenishment. Group correction restores most
validity, while both factors are needed for
the best FCD in this model.
\begin{table}[!htbp]
\caption{\textbf{Both within-group and between-group corrections contribute.} Group and identity corrections are varied during both generation and assembly in a fixed trained model.}
\label{tab:head}
\centering
\medskip
\begingroup
\small
\setlength{\tabcolsep}{3pt}
\begin{tabular*}{\linewidth}{@{\extracolsep{\fill}}lrrr@{}}
\toprule
Correction & Valid / attempted & Validity (\%) & FCD $\downarrow$ \\
\midrule
Neither & 3,313 / 4,096 & 80.88 & 8.2378 \\
Group only & 4,047 / 4,096 & 98.80 & 4.0742 \\
Identity only & 3,266 / 4,096 & 79.74 & 4.1570 \\
Both & 4,090 / 4,096 & 99.85 & 0.6264 \\
\bottomrule
\end{tabular*}
\par\smallskip
\begin{minipage}{\linewidth}\footnotesize
All four arms score the first 3,000 valid outputs from 4,096 attempts against the same 11,228-molecule internal reference. This common evaluation size is supported by all four conditions, without replenishment. The model uses the 38,951-entry vocabulary and the replication-model training protocol in Appendix~\ref{app:protocols}.
\end{minipage}
\endgroup
\end{table}

\paragraph{Predictive information beyond chemical features.}
We test whether identity corrections provide predictive information outside
the score space spanned by the chemical features, using the projector in
Appendix~\ref{app:mechanism-algebra}. At $t=0.5$, we compare the full predictor
with predictors retaining only the inside-span or outside-span correction.
Target bits per site measure the base-two log probability assigned to the
correct identity at each node; a loss in bits denotes a reduction in that
probability relative to the full predictor.

In the final aligned checkpoint, removing the outside-span correction loses
1.79 target bits/site (paired molecule-bootstrap 95\% interval [1.68, 1.89])
and 27.8 top-1 accuracy points. Removing the inside-span component loses
1.36 [1.14, 1.60] bits/site. Both components contribute in the trained model.
The outside-span loss remains 1.79 bits on sites with unique feature rows,
showing that the effect extends beyond feature collisions. The probe uses
512 validation molecules excluding exact training identities, with 2,730
sites whose target identities occur in training; 2,669 of those sites have
unique feature rows. The full predictor correctly classifies none of the
52 targets whose identities are absent from training.

\paragraph{Effect of chemical alignment on prediction.}
We compare independently trained aligned and shuffled predictors on the same
corrupted inputs at $t=0.5$, using final EMA weights. Gain is the aligned
predictor's increase in target log probability, in bits per site; its
group-level component measures the increase in log probability of the correct
attachment group. The probe covers 3,036 validation molecules excluded from
checkpoint selection and 16,331 sites. After excluding 36 exact training
molecules, the gain for training-observed target identities is 0.0107
[$-0.0027$, 0.0244] bits/site. Before this exclusion, the corresponding
group-level gain is 0.0119 [0.0037, 0.0198]. In that original population,
targets with identities absent from training gain 12.15 bits but reach only
2.91\% top-1 accuracy versus 0\% for the shuffled predictor.

The difference in mean gain between rare identities (1--10 training
occurrences) and common identities (more than 1,000) is inconclusive.
Intervals use 2,000 paired molecule-bootstrap resamples from one trained
pair, without multiplicity adjustment.
\FloatBarrier

\section{Transfer to MOSES}
\label{app:moses-training}
\subsection{Matched training and evaluation}
The fixed design is trained on 1,505,430 training examples and 79,233 internal validation examples, using a 51,879-entry vocabulary and 8,192 validation examples
for checkpoint selection. Three
matched training runs each train aligned, shuffled and conventional output
predictors for 42 epochs (493,962 updates), sharing the split, backbone
initialization, source priors, corruption, EMA selection and assembly recipe.
Aligned and shuffled models have approximately 30.8M parameters. Shuffling permutes whole
output feature rows within positive/zero-count strata. The conventional head
is a $384\!\to\!262\!\to\!51{,}879$ ReLU MLP with biases and the same fixed
log-frequency offset, matching the parameter count within 0.05\%.

Each checkpoint produces 30,720 raw attempts at 250 and 1000 steps, with
paired canonical/prior assembly. The primary population is the first 25,000
raw attempts, prior assembly, scored against all 176,225 official scaffold-test
(TestSF) molecules \citep{moses}. Failures are not replenished. The official
176,074-molecule Test reference, samples of 30,000 attempted and 25,000 valid outputs, canonical
assembly and 1000 steps are separate sensitivity analyses. Novelty uses unique valid
canonical outputs against official training strings. Validation-selected and
final EMA checkpoints are reported separately.

\subsection{Results across training runs and checkpoint policies}
Table~\ref{tab:moses-matched-complete} gives every run's primary FCD under
both checkpoint policies and both step counts. At 250 steps, selected aligned
FCD averages 0.5349 versus 0.5507 shuffled and 0.5589 conventional, reductions
of 2.9\% and 4.3\%. The common-final means are 0.5393, 0.5483 and 0.5465.
The mean FCD advantage also appears on MOSES, with a smaller effect than on ZINC.
\begin{table}[!htbp]
\caption{\textbf{MOSES: three training runs and both checkpoint policies.} FCD against TestSF; smaller is better.}
\label{tab:moses-matched-complete}
\centering
\medskip
\begingroup
\small
\setlength{\tabcolsep}{3pt}
\begin{tabular*}{\linewidth}{@{\extracolsep{\fill}}llrrrr@{}}
\toprule
Head & Run & \multicolumn{2}{c}{250 steps} & \multicolumn{2}{c}{1000 steps} \\
& & Selected & Final & Selected & Final \\
\midrule
Aligned & A & 0.5282 & 0.5443 & 0.5169 & 0.5245 \\
Aligned & B & 0.5622 & 0.5622 & 0.5299 & 0.5299 \\
Aligned & C & 0.5142 & 0.5115 & 0.5333 & 0.5251 \\
Mean $\pm$ SD &  & $0.5349\pm0.0247$ & $0.5393\pm0.0257$ & $0.5267\pm0.0087$ & $0.5265\pm0.0030$ \\
\midrule
Shuffled & A & 0.5626 & 0.5504 & 0.5401 & 0.5443 \\
Shuffled & B & 0.5590 & 0.5536 & 0.5435 & 0.5539 \\
Shuffled & C & 0.5306 & 0.5408 & 0.5355 & 0.5237 \\
Mean $\pm$ SD &  & $0.5507\pm0.0175$ & $0.5483\pm0.0066$ & $0.5397\pm0.0040$ & $0.5407\pm0.0155$ \\
\midrule
Conventional & A & 0.5700 & 0.5422 & 0.5586 & 0.5437 \\
Conventional & B & 0.5475 & 0.5544 & 0.5324 & 0.5325 \\
Conventional & C & 0.5591 & 0.5430 & 0.5317 & 0.5445 \\
Mean $\pm$ SD &  & $0.5589\pm0.0112$ & $0.5465\pm0.0068$ & $0.5409\pm0.0153$ & $0.5402\pm0.0067$ \\
\bottomrule
\end{tabular*}
\par\smallskip
\begin{minipage}{\linewidth}\footnotesize
Each cell uses valid outputs among the first 25,000 raw attempts, without refill, prior assembly and the 176,225-molecule scaffold-held-out TestSF reference. Selected and final EMA policies are not pooled. Means and sample standard deviations are across independent training seeds. All heads use the matched recipe and approximately 30.8M parameters.
\end{minipage}
\endgroup
\end{table}

At 250 steps, aligned prediction wins two of three selected pairs against each
comparator. At final EMA it wins two of three against shuffling and one of
three against the conventional head. At 1000 steps it wins three of three
final pairs against the conventional head and two of three against shuffling.
Both comparators retain higher primary novelty. These reversals distinguish a
positive mean effect from a uniform improvement across training runs.

Eight 25,000-molecule training draws score scaffold-test FCD
$0.518\pm0.012$ (mean and sample SD), providing the training-data floor for
this evaluation.
\FloatBarrier

\section{Property-conditioned generation}
\label{app:conditional-complete}
\subsection{Adaptation and baseline training}

\paragraph{\model{} adaptation.}
The conditional split contains 207,114 unique training identities and 10,800 validation
identities. The property vector holds training-standardized QED \citep{qed} and logP together
with a two-entry indicator naming the requested property; the unrequested
property's standardized value and its indicator are both zero. For each of the
12 frozen transformer layers, a separate $4\!\rightarrow\!64\!\rightarrow\!544$
MLP with SiLU activation maps this four-dimensional input to offsets of widths
384/96/64 for the node, edge and graph states, broadcast across valid nodes and
edges after that layer's output. The final adapter projection starts at zero and
a missing condition multiplies every offset by zero, so the unconditioned
computation is preserved exactly. Conditioning changes neither the categorical
vocabulary nor the linear probability path.

Training updates only the approximately 0.43M adapter parameters. The pretrained backbone and
the output predictor stay frozen and in evaluation mode. Each real and shuffled
arm uses matched initial adapters and matched data and corruption streams for
20,000 updates at batch 128 of AdamW with learning rate $2\times10^{-4}$, weight decay
$10^{-12}$ and EMA decay 0.999, retaining the molecular endpoint cross-entropy
objective. Each training example requests one property at random, with 10\%
condition dropout. Every 2,000 updates, a fixed 2,048-molecule validation subset
selects the adapter by denoising loss. Three real-label adaptations share one frozen backbone; two have matched fixed-label-permutation controls.
Generation multiplies the learned offsets by a selected strength in $\{1,2\}$;
this is adapter scaling, not classifier-free extrapolation between two separately
evaluated predictions.

\paragraph{External conditional models.}
FragFM uses its native noisy property-predictor objective and guidance, with the
molecular generator trained on the shared benchmark data frozen. Separate
6.57M-parameter QED and logP predictors train for 50,000 updates at batch 256 of AdamW with
learning rate $5\times10^{-4}$, a 10,000-update warmup and weight decay
$10^{-12}$, using the authors' four-layer fragment and graph predictors; fixed
validation MSE selects the EMA checkpoints. Graph-only and graph-plus-bag
guidance remain separate regimes, both retaining 500 generation steps. Graph DiT
trains a separate 157.9M-parameter model per property from scratch: width 1152,
six layers, 16 attention heads, 500 diffusion steps, Adam with learning rate
$2\times10^{-4}$, and 100 epochs (161,900 updates) at effective batch 128 and condition-drop probability
0.1, with fixed validation noising on 2,048 molecules selecting the checkpoint.
Training labels and generation requests use the same training-set standardization. The
adaptation-based and from-scratch methods therefore differ in training cost and
objective.

\paragraph{Guidance selection.}
For each property and candidate strength, validation generation drew 512 attempts
at each of the three targets, and the setting maximizing their mean raw hit rate
was selected, with ties resolved toward the first candidate. Guidance was fixed
before final evaluation. Table~\ref{tab:guidance-settings} gives the candidate and selected strengths.
The selection made on the real \model{} arm was applied unchanged to its
shuffled-label control.

\begin{table}[!htbp]
\caption{\textbf{Guidance candidates and validation-selected strengths.}
FragFM pairs list graph and bag strengths; graph-only guidance uses the graph entry.}
\label{tab:guidance-settings}
\centering\small
\begin{tabular*}{\linewidth}{@{\extracolsep{\fill}}llll@{}}
\toprule
Method & Property & Candidates & Selected\\
\midrule
\model{} & QED, logP & 1; 2 & 2\\
Graph DiT & QED, logP & 1; 2 & 2\\
FragFM & QED & (20, 1); (1042, 52.09) & (1042, 52.09)\\
FragFM & logP & (20, 1); (9.814, 0.491) & (20, 1)\\
\bottomrule
\end{tabular*}
\end{table}

\subsection{Targets, metrics and matched timing}
Targets are the 10th, 50th and 90th training quantiles (q10, q50 and q90):
approximately 0.529, 0.763 and 0.885 for QED, and 0.571, 2.626 and 4.186 for
logP. All target tolerances and guidance choices were fixed before final
scoring. Each external method uses one trained model per property. A hit is a valid output whose property lies
within $\pm0.05$ QED or $\pm0.5$ logP of the requested value, endpoints included.
Raw hit rates divide hits by all 4,096 attempts, so invalid attempts and repeated
hits stay in the denominator and in the cost accounting. Distinct novel hits
count each hitting molecular identity once and additionally require it to be
absent from the conditional training identities; identity ignores stereochemistry,
and all methods use the same validity, identity and property checks.

Target-window FCD uses the first 1,000 valid outputs from each eligible cell,
including valid misses, rather than a subset selected for hitting the target. It
scores them against the first 1,000 held-out reference identities inside the same
target window; generated duplicates remain and there is no refill. Conditional identity and
reference preprocessing differ from the unconditional protocol, so these values
are not numerically comparable with the unconditional ZINC evaluation.

\paragraph{Matched conditional timing protocol.}
\label{app:conditional-timing}
The matched comparison measures every primary 500-step setting: three targets
for each property, three \model{} adaptations, both FragFM guidance regimes and
Graph DiT. Checkpoints and validation-selected guidance strengths are those used
for the evaluation samples. Each setting has three 256-attempt timing repetitions, with method order rotated across repetitions and each setting run
in an isolated process with its model's software versions at batch 128 after 128 warmup attempts, with four CPU threads. Within a
target comparison, all methods run on the same physical GPU.

The primary timer starts before random-state and graph initialization and ends
after generation, decoding and the shared property, identity, novelty and hit
checks. Guidance, failed assemblies and invalid outputs remain inside this
interval. A second endpoint before annotation records generation and decoding
alone. Loading, warmup, file writing, FCD scoring, training and guidance search
are excluded for every method, and GPU work is synchronized at the generation
boundaries. FragFM timing uses its strict assembly decoder.

For each adaptation or baseline setting $s$, let $U_s$ be the number of distinct
novel target hits in its 4,096-attempt evaluation sample, and let
$t_{s,r}$ be the annotation-inclusive time for timing repetition $r$. The
estimated useful rate is
\begin{equation}
q_s=\frac{U_s}{4{,}096},\qquad
\widehat r_s=\frac{256q_s}{\operatorname{median}_{r=1,2,3}t_{s,r}}.
\label{eq:conditional-rate}
\end{equation}
The reported \model{} value is the arithmetic mean of its three adaptation-specific rates. Distinct-hit yield depends on the evaluation sample size.

\subsection{Conditional yield, throughput and fidelity}
\label{app:cond-grids}
Table~\ref{tab:app-cond-full} reports all targets, models and guidance settings
across five metrics.
Both FragFM settings reach every target more often per attempt than \model{}
does; at the high-QED target FragFM also attains the lower target-window FCD.

Figure~\ref{fig:figure5-supp} separates hit rate from throughput. \model{} completes
21.3--25.3 times as many attempts per second as either FragFM setting and
12.3--13.0 times as many as the fastest baseline at each target. That
throughput more than offsets a per-attempt hit rate between 0.34 and 0.96 of
FragFM's, depending on the target: the gap is widest at low logP and nearly
closed at high QED. The resulting rate advantage is 7.8--20.8-fold against the
strongest baseline at each target, 7.8--22.5-fold against either FragFM setting
and 11.4--24.4-fold against Graph DiT.

\begin{table*}[t]
\caption{\textbf{Complete property-targeted generation results.} Distinct novel hits per second, hit rate, distinct novel hits per attempt, target-window FCD and validity for every target, model and guidance setting.}
\label{tab:app-cond-full}
\centering
\medskip
\begingroup
\small
\setlength{\tabcolsep}{3pt}
\begin{tabular*}{\linewidth}{@{\extracolsep{\fill}}lrrrrr@{}}
\toprule
Model & Distinct novel & Hit rate & Distinct novel & Target-window & Valid \\
& hits / second & (\%) & hits / attempt & FCD $\downarrow$ & (\%) \\
\midrule
\multicolumn{6}{@{}l}{\emph{QED, low (10th) quantile}: target 0.529 $\pm$ 0.05}\\[1pt]
\hspace{6pt}\model{} & \textbf{2.719} & 18.12 & 0.1739 & 5.156 & 99.71 \\
\hspace{6pt}FragFM, graph guidance & 0.236 & 34.06 & 0.3386 & 5.330 & 90.92 \\
\hspace{6pt}FragFM, graph + bag guidance & 0.225 & 35.45 & 0.3503 & 5.346 & 90.21 \\
\hspace{6pt}Graph DiT & 0.119 & 9.69 & 0.0969 & 11.614 & 43.65 \\
\addlinespace[2pt]
\multicolumn{6}{@{}l}{\emph{QED, median (50th) quantile}: target 0.763 $\pm$ 0.05}\\[1pt]
\hspace{6pt}\model{} & \textbf{4.871} & 31.63 & 0.3089 & 2.927 & 99.66 \\
\hspace{6pt}FragFM, graph guidance & 0.376 & 51.88 & 0.5159 & 3.434 & 97.51 \\
\hspace{6pt}FragFM, graph + bag guidance & 0.376 & 51.46 & 0.5107 & 3.674 & 97.34 \\
\hspace{6pt}Graph DiT & 0.273 & 21.85 & 0.2185 & 5.307 & 57.30 \\
\addlinespace[2pt]
\multicolumn{6}{@{}l}{\emph{QED, high (90th) quantile}: target 0.885 $\pm$ 0.05}\\[1pt]
\hspace{6pt}\model{} & \textbf{8.430} & 52.99 & 0.5224 & 3.465 & 99.72 \\
\hspace{6pt}FragFM, graph guidance & 0.375 & 55.15 & 0.5479 & 3.284 & 96.56 \\
\hspace{6pt}FragFM, graph + bag guidance & 0.404 & 59.08 & 0.5874 & 3.256 & 98.44 \\
\hspace{6pt}Graph DiT & 0.345 & 27.81 & 0.2778 & 4.678 & 63.09 \\
\addlinespace[2pt]
\multicolumn{6}{@{}l}{\emph{logP, low (10th) quantile}: target 0.571 $\pm$ 0.5}\\[1pt]
\hspace{6pt}\model{} & \textbf{3.519} & 21.96 & 0.2145 & 3.518 & 99.58 \\
\hspace{6pt}FragFM, graph guidance & 0.453 & 64.23 & 0.6392 & 3.636 & 93.24 \\
\hspace{6pt}FragFM, graph + bag guidance & 0.393 & 60.94 & 0.6047 & 4.129 & 91.11 \\
\hspace{6pt}Graph DiT & 0.175 & 13.87 & 0.1387 & 7.231 & 75.66 \\
\addlinespace[2pt]
\multicolumn{6}{@{}l}{\emph{logP, median (50th) quantile}: target 2.626 $\pm$ 0.5}\\[1pt]
\hspace{6pt}\model{} & \textbf{4.784} & 31.48 & 0.3050 & 3.079 & 99.67 \\
\hspace{6pt}FragFM, graph guidance & 0.519 & 71.19 & 0.7068 & 4.873 & 96.34 \\
\hspace{6pt}FragFM, graph + bag guidance & 0.540 & 73.71 & 0.7327 & 4.731 & 97.02 \\
\hspace{6pt}Graph DiT & 0.398 & 31.25 & 0.3125 & 6.412 & 64.62 \\
\addlinespace[2pt]
\multicolumn{6}{@{}l}{\emph{logP, high (90th) quantile}: target 4.186 $\pm$ 0.5}\\[1pt]
\hspace{6pt}\model{} & \textbf{5.277} & 33.57 & 0.3288 & 4.040 & 99.72 \\
\hspace{6pt}FragFM, graph guidance & 0.451 & 63.62 & 0.6348 & 5.385 & 96.73 \\
\hspace{6pt}FragFM, graph + bag guidance & 0.435 & 66.60 & 0.6616 & 4.926 & 97.39 \\
\hspace{6pt}Graph DiT & 0.465 & 36.79 & 0.3677 & 8.605 & 58.98 \\
\bottomrule
\end{tabular*}
\par\smallskip
\begin{minipage}{\linewidth}\footnotesize
All settings use 500 steps, 4,096 attempts and validation-selected guidance. \model{} rows average three adaptations of one frozen backbone; each baseline regime is one trained model. A hit is valid and inside the target window; distinct novel hits additionally count each non-training identity once. Rates combine each population's useful yield with separate timing repetitions (Equation~\ref{eq:conditional-rate}). Target-window FCD scores the first 1,000 valid outputs, including misses, against 1,000 held-out in-window references.
\end{minipage}
\endgroup
\end{table*}

\begin{figure}[!htbp]
\centering
\includegraphics[width=\linewidth]{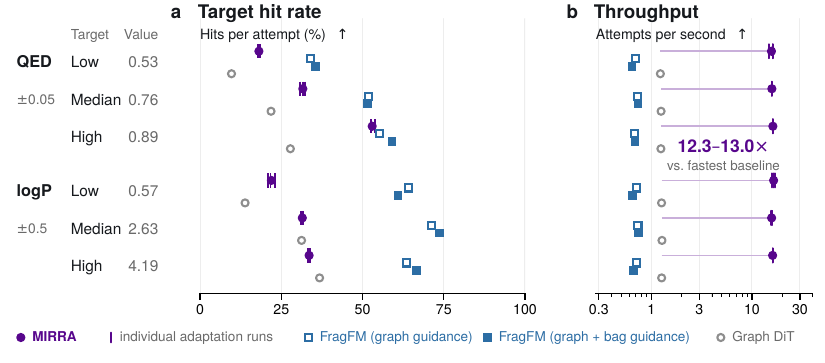}
\input{figures/figure5_supp_caption}
\end{figure}

\subsection{Label controls and adaptation variability}
Table~\ref{tab:app-cond-controls} reports the two runs with matched real-label,
shuffled-label and unconditional populations. Real conditioning increases mean
hit rate at all six targets relative to both controls, with smaller gains at
the central targets. Each entry averages two matched 4,096-attempt populations. Filtering
an unconditional population within the same attempt budget cannot increase its
number of hits; it only discards misses. Thus the unconditional column also
records the raw-budget hit yield of sample-and-filter.

The right-hand columns show each of the three primary adaptations' useful rates,
computed separately by Equation~\ref{eq:conditional-rate}. The rates describe adaptation and sampling variability on one frozen backbone.
\begin{table}[!htbp]
\caption{\textbf{Condition-label controls and individual adaptation rates.} Hit-rate controls average two matched adapter runs; distinct novel hit rates are reported for each of three adapter runs.}
\label{tab:app-cond-controls}
\centering
\medskip
\begingroup
\small
\setlength{\tabcolsep}{3pt}
\begin{tabular*}{\linewidth}{@{\extracolsep{\fill}}lrrrrrr@{}}
\toprule
Target & \multicolumn{3}{c}{Hit rate (\%), two-run mean} & \multicolumn{3}{c}{Distinct novel hits/s, real labels} \\
& Real & Shuffled & Uncond. & Run A & Run B & Run C \\
\midrule
QED, low & 17.91 & 8.01 & 8.29 & 2.540 & 2.748 & 2.869 \\
QED, median & 31.56 & 28.50 & 29.36 & 4.956 & 4.828 & 4.830 \\
QED, high & 53.14 & 26.09 & 25.67 & 8.503 & 8.471 & 8.317 \\
logP, low & 21.36 & 11.01 & 10.41 & 3.494 & 3.452 & 3.610 \\
logP, median & 31.46 & 27.98 & 26.40 & 4.914 & 4.765 & 4.673 \\
logP, high & 33.70 & 15.41 & 15.41 & 5.264 & 5.311 & 5.255 \\
\bottomrule
\end{tabular*}
\par\smallskip
\begin{minipage}{\linewidth}\footnotesize
Left: matched runs A and B, 500 steps and 4,096 attempts per cell. Shuffling permutes condition labels during adapter training; unconditional generation uses the same backbone and attempt budget. Right: useful rates for all three real-label adaptations, each computed using its own quality population and three timing repetitions. Rates are estimates under Equation~\ref{eq:conditional-rate}.
\end{minipage}
\endgroup
\end{table}

\newpage
\begingroup
\setlength{\parskip}{4pt}
\setlength{\intextsep}{8pt plus 2pt minus 2pt}
\setlength{\textfloatsep}{10pt plus 2pt minus 2pt}
\section{Polymer generation}
\label{app:polymer-training}

\subsection{Data, representations and baseline adaptations}
We use the Open Macromolecular Genome authors' selected all-reactions training
set of 99,500 source products across 17 reaction labels rather than the
complete OMG database.\footnote{\url{https://github.com/TheJacksonLab/OpenMacromolecularGenome/tree/e4681336c9c7aa1b62ea3f4e31dcf60154d4ba5e}.}
Canonicalization identifies 99,476 distinct periodic objects. Rotation,
reversal and repeated-unit aliases share one split group, so no reserved
identity can reappear in training under a relabeling. The split holds 79,580
training, 9,947 validation and 9,949 reserved test objects. Internal
generation controls use validation only, including 996 objects with unseen
blocks. Checkpoints and generated samples are fixed before evaluation on the 9,949 held-out test objects
(Table~\ref{tab:short-polymer}). This experiment evaluates structural generation
on a known linear chain topology.

The comparison uses two 4,096-attempt draws per model and setting. MIRRA and Graph DiT use 500 sampling steps, with MIRRA also evaluated at 250 steps from the same checkpoint. The VAE uses native autoregressive decoding. Validity is computed over all 4,096 attempts in each sampling run; structural distances condition on scorable outputs, whose fractions are reported separately.

\paragraph{Representations and information supplied to each method.}
MIRRA predicts oriented building blocks on a supplied linear chain, using
circular attention and reversal-compatible features; its 14.5M-parameter model
trains for 200,000 updates and uses the validation-selected checkpoint. The
external models instead generate complete repeat-unit encodings: OMG-VAE emits
SELFIES strings \citep{selfies}, and Graph DiT emits atom and bond labels including wildcard
attachment atoms. Neither receives MIRRA's fixed chain connectivity. The
external comparison therefore measures complete modeling approaches under a
common split and scorer, including the advantage of a task-specific
representation. The paired internal controls reported below use the same
representation in both arms.

\paragraph{OMG-VAE adaptation.}
The 3.95M-parameter model retains the OMG authors' CNN encoder,
152-dimensional latent representation and four-layer bidirectional GRU decoder
called one token at a time, with an auxiliary network predicting standardized
logP normalized by backbone atom count. Training uses the authors'
reconstruction and IMQ-MMD objective with MMD weight 4.345, Adam learning rate
0.002, weight decay $10^{-5}$, batch 32 and 60 epochs (149,220 updates); the
learning rate falls tenfold at 40 epochs. Autoregressive reconstruction over
all 9,947 validation objects under fixed latent draws selects the checkpoint,
epoch 45. The SELFIES alphabet and maximum length come from training data
only, and the fixed initial attachment token is removed for learning and
prepended on decode. Generation samples a Gaussian latent prior and selects tokens greedily. Decoding truncates before the first no-operation token or the second generated attachment token, with the initial attachment token prepended afterward. Outputs are prior samples rather than reconstructions or optimized latent vectors.

\paragraph{Graph DiT adaptation.}
The 160.1M-parameter model retains six transformer layers of width 1152 with
16 heads, 500 diffusion steps and Adam learning rate $2\times10^{-4}$. It
trains from scratch for 100 epochs (62,200 optimizer updates) at effective
batch 128. Wildcard attachment atoms are included as nodes when estimating graph sizes and no-edge frequencies. Graph DiT uses normalized time during training and sampling. Fixed noising
on up to 2,048 validation objects selects the checkpoint, epoch 100. The
maximum graph size in the training set defines model support.
Generation uses the trained size prior and predicts the complete graph without
requesting any property.

\paragraph{Periodic validity.}
Ordinary chemical validity requires only a chemically parseable output.
Periodic validity is stricter: the repeat unit must be connected, must carry
the two required attachment ports, and must canonicalize periodically, with
rotation, reversal and repeated-unit aliases collapsing to a single identity.
Valid baseline outputs containing building blocks absent from MIRRA's
vocabulary are accepted. A periodically valid output is scorable if its
seven-copy chain can be assembled and sanitized and its central-unit fingerprint
and all six structural descriptors can be computed. Structural MMD and
descriptor distances use these scorable outputs; validity and the scorable
fraction are reported separately over all attempts.

\subsection{Structural metrics and internal controls}
The aligned and permuted models are sampled at 250 sampling steps (primary) and 1000 steps
(sensitivity) with 4,096 raw attempts per population, no rejection or
replenishment, matched requested lengths, the selected checkpoints and the
full 41,266-token training vocabulary. The reference is the complete
9,947-object validation population, all scorable.

Structural features are $\ell_2$-normalized 2,048-dimensional Morgan counts
(radius two, chirality enabled), rooted at real atoms of the central unit in a
seven-copy chain. For feature vectors $x,y$, the kernel averages
$\exp[-\|x-y\|^2/(2s^2f)]$ over $f\in\{0.5,1,2\}$, with $s^2=1.158$ fixed as
the median squared distance of 4,096 unequal-index pairs drawn from 2,048
distinct training objects. The unbiased MMD$^2$ estimator \citep{mmd} excludes
within-population self-pairs and retains duplicates, so it may be negative.
The main external table reports MMD$^2$ in units of $10^{-3}$; the smaller
internal differences below use $10^{-5}$. Descriptor
distances are Wasserstein distances over primitive-unit mass, central topological polar surface area (TPSA),
and heteroatom, aromatic-heavy, branched-heavy and nonring-single-heavy-bond
fractions, each divided by its training-calibration sample standard deviation.
These are structural descriptors, not measured bulk polymer properties.

\paragraph{Polymer prediction with shared outputs.}
Both arms have approximately 14.52M parameters and train for 200,000 updates
from matching initial weights on the same data. Meaningful output sharing lowers
the checkpoint-selection objective throughout training, with a 2.9\% reduction
at the final update. Selection uses mean endpoint cross-entropy on 1,024
validation objects, averaging each object's token-average loss so that long
objects receive no greater weight. Table~\ref{tab:polymer-selection} summarizes
the selected and final checkpoints from this trained pair.

\begin{table}[!htbp]
\caption{\textbf{Chemically aligned output sharing lowers polymer prediction loss.}
Object-mean endpoint cross-entropy in nats.}
\label{tab:polymer-selection}
\centering\small
\begin{tabular*}{\linewidth}{@{\extracolsep{\fill}}lrrr@{}}
\toprule
Output features & Selected update & Selected loss & Final loss\\
\midrule
Aligned & 192,000 & 2.4617 & 2.4632\\
Permuted & 198,000 & 2.5347 & 2.5365\\
\bottomrule
\end{tabular*}
\end{table}

At 250 steps the permuted-minus-aligned MMD$^2$ difference is 4.96, and
aligned sharing improves five of six descriptor distances; the branched-heavy
fraction favors the permuted control (aligned: 0.0359; permuted: 0.0153). Both arms are
fully valid, with 98.29\% and 97.66\% novel draws, whereas a training-resampling
control reaches comparably low MMD$^2$ with zero novelty. At 1000 steps the
ordering reverses: aligned MMD$^2$ is 4.88 against permuted 3.60, aligned
sharing is closer only on TPSA and heteroatom fraction, and novelty is 98.36\%
and 97.56\%. The favorable primary point estimate therefore reverses at 1000 steps.

\paragraph{Additional sampling runs at 250 steps.}
Four further paired draws repeat the 250-step comparison with 4,096 raw
attempts per arm, fixed requested lengths and the same 9,947-object reference
(Table~\ref{tab:app-polymer-repeats}). Three of four paired differences favor
aligned sharing, and the mean permuted-minus-aligned difference is
$0.97\pm1.94$ in units of $10^{-5}$ (sample standard deviation across draws). Five of six mean
descriptor distances favor aligned sharing, with the branched-heavy fraction
again favoring the control (aligned: 0.0266; permuted: 0.0227). Mean novelty is 98.55\%
against 97.56\%, favoring aligned sharing in every paired draw, and both arms
retain 100\% validity. These additional draws use the same fixed checkpoints.

\begin{table}[!htbp]
\caption{\textbf{Aligned output sharing lowers polymer structural MMD$^2$ in
three of four additional 250-step draws.} Paired draws of 4,096 attempts per arm
against the full validation reference.}
\label{tab:app-polymer-repeats}
\centering
\begingroup
\small
\setlength{\tabcolsep}{3pt}
\begin{tabular*}{\linewidth}{@{\extracolsep{\fill}}lrrrrr@{}}
\toprule
Draw & Aligned & Permuted & Difference & Novelty A (\%) & Novelty P (\%) \\
\midrule
1 & 2.133 & 2.750 & 0.62 & 98.63 & 97.27 \\
2 & 5.900 & 6.749 & 0.85 & 98.54 & 97.61 \\
3 & 8.946 & 7.810 & $-1.14$ & 98.76 & 97.97 \\
4 & 5.193 & 8.755 & 3.56 & 98.27 & 97.41 \\
\midrule
Mean & 5.543 & 6.516 & 0.97 & 98.55 & 97.56 \\
\bottomrule
\end{tabular*}
\par\smallskip
\begin{minipage}{\linewidth}\footnotesize
Unbiased MMD$^2$ in units of $10^{-5}$; positive differences (permuted minus
aligned) favor aligned sharing. Novelty A and P are the aligned and permuted
novel fractions. Validity is 100\% and uniqueness at least 99.85\% in every
population.
\end{minipage}
\endgroup
\end{table}

\subsection{Conventional output head on polymers}
\label{app:polymer-head-control}

A matched comparison replaces only the output head in the circular backbone.
The conventional head is a $256\!\to\!263\!\to\!41{,}266$ ReLU MLP, plus the
same fixed log-frequency prior; its hidden width was chosen by nearest
parameter count before training, matching the shared-output model to within 0.03\%, at approximately 14.52M parameters each. Both arms retain structural
inputs, full categorical support, the same 200,000-update recipe, shared
initial backbone weights and identical training streams. The
validation-selected EMA is primary and the final-update EMA is a separate
checkpoint sensitivity. Each head and
policy contributes two paired 250-step draws of 4,096 attempts against the same
9,947-object reference. This comparison uses a separate trained model pair.

\begin{table}[!htbp]
\caption{\textbf{The shared output head achieves lower structural discrepancy
than a capacity-matched conventional head under both checkpoint policies.} Paired 250-step draws of
4,096 attempts each; unbiased structural MMD$^2$ in units of $10^{-5}$.}
\label{tab:app-polymer-head}
\centering
\begingroup
\small
\setlength{\tabcolsep}{3pt}
\begin{tabular*}{\linewidth}{@{\extracolsep{\fill}}llrr@{}}
\toprule
Checkpoint policy & Draw & Aligned head & Conventional head \\
\midrule
Validation-selected EMA & 1 & 3.243 & 11.36 \\
& 2 & 0.5184 & 9.745 \\
& Mean & 1.881 & 10.55 \\
\midrule
Final-update EMA & 1 & 3.163 & 13.84 \\
& 2 & 1.081 & 11.19 \\
& Mean & 2.122 & 12.51 \\
\bottomrule
\end{tabular*}
\par\smallskip
\begin{minipage}{\linewidth}\footnotesize
Values summarize sampling variability across two draws per fixed checkpoint. Final-update descriptor distances (aligned/conventional): heteroatom fraction, 0.0332/0.0220; nonring-single-heavy-bond fraction, 0.0303/0.0271.
\end{minipage}
\endgroup
\end{table}

Aligned sharing has the lower MMD$^2$ in all four paired draws
(Table~\ref{tab:app-polymer-head}). The
conventional-minus-aligned difference averages $8.671\pm0.786$ under the
validation-selected policy and $10.39\pm0.400$ under the final-update policy
(mean $\pm$ draw standard deviation). Under the validation-selected policy
aligned sharing is closer on all six descriptors; under the final-update
policy it is closer on four, with heteroatom and nonring-single-heavy-bond fractions favoring the conventional head. Validity is 100\% in all eight populations and
novelty ranges from 98.24\% to 99.15\%.

\FloatBarrier

\endgroup

\end{document}